\documentclass[10pt,journal,compsoc]{IEEEtran}

\ifCLASSOPTIONcompsoc
  \usepackage[nocompress]{cite}
\else
  \usepackage{cite}
\fi

\ifCLASSINFOpdf
\else
  Online Image Vectorizern (dvipsone, dvipdf, if not using dvips). graphicx
\fi

\usepackage{amsmath,amsfonts}
\usepackage{array}
\usepackage[caption=false,font=normalsize,labelfont=sf,textfont=sf]{subfig}
\usepackage{textcomp}
\usepackage{stfloats}
\usepackage{url}
\usepackage{verbatim}
\usepackage{cite}
\usepackage{graphicx}
\usepackage{tabularx}
\usepackage{bm}
\usepackage{algorithm,algpseudocode}
\usepackage{xcolor,colortbl}
\usepackage{color, colortbl}

\usepackage{times}
\usepackage{epsfig}
\usepackage{amssymb}

\usepackage{pifont}
\usepackage{xcolor, colortbl}
\usepackage{enumitem}
\usepackage{multirow}
\usepackage{siunitx}
\usepackage{xspace}
\usepackage{booktabs}

\usepackage[pagebackref=true,breaklinks=true,letterpaper=true,colorlinks,bookmarks=false]{hyperref}
\usepackage[capitalize]{cleveref}
\usepackage{orcidlink}

\usepackage{pifont}

\crefname{section}{Sec.}{Secs.}
\Crefname{section}{Section}{Sections}
\Crefname{table}{Table}{Tables}
\crefname{table}{Tab.}{Tabs.}

\definecolor{GRAY}{gray}{0.85}
\newcommand{\name}{SpikeMOT\xspace}
\newcommand{\splittab}[2]{ \begin{tabular}{@{}c@{}} {#1} \\ {#2} \end{tabular} }

\begin{document}

\title{SpikeMOT: Event-based Multi-Object Tracking with Sparse Motion Features}

\author{
Song Wang\textsuperscript{*}~\orcidlink{0000-0002-1813-5865}, Zhu Wang\textsuperscript{*}~\orcidlink{0000-0002-3859-1008}, Can Li~\orcidlink{0000-0003-3795-2008}, Xiaojuan Qi~\orcidlink{0000-0002-4285-1626}, Hayden Kwok-Hay So~\orcidlink{0000-0002-6514-0237},~\IEEEmembership{Senior Member,~IEEE}
\IEEEcompsocitemizethanks{\IEEEcompsocthanksitem Song Wang, Zhu Wang, Can Li, Xiaojuan Qi, Hayden Kwok-Hay So are with the Department of Electrical and Electronic Engineering at The University of Hong Kong, Hong Kong, China.
\IEEEcompsocthanksitem {Email: wangsong, zhuwang@connect.hku.hk; lican, xjqi, hso@eee.hku.hk}
}
}

\IEEEtitleabstractindextext{
\begin{abstract}
In comparison to conventional RGB cameras, the superior temporal resolution of event cameras allows them to capture rich information between frames, making them prime candidates for object tracking. Yet in practice, despite their theoretical advantages, the body of work on event-based multi-object tracking (MOT) remains in its infancy, especially in real-world settings where events from complex background and camera motion can easily obscure the true target motion. In this work, an event-based multi-object tracker, called SpikeMOT, is presented to address these challenges. SpikeMOT leverages spiking neural networks to extract sparse spatiotemporal features from event streams associated with objects. The resulting spike train representations are used to track the object movement at high frequency, while a simultaneous object detector provides updated spatial information of these objects at an equivalent frame rate. To evaluate the effectiveness of SpikeMOT, we introduce DSEC-MOT, the first large-scale event-based MOT benchmark incorporating fine-grained annotations for objects experiencing severe occlusions, frequent trajectory intersections, and long-term re-identification in real-world contexts. Extensive experiments employing DSEC-MOT and another event-based dataset, named FE240hz, demonstrate SpikeMOT's capability to achieve high tracking accuracy amidst challenging real-world scenarios, advancing the state-of-the-art in event-based multi-object tracking.

\end{abstract}

}

\maketitle

\begingroup\renewcommand\thefootnote{*}
\footnotetext{Equal contribution}
\endgroup

\IEEEdisplaynontitleabstractindextext

%
\IEEEpeerreviewmaketitle

\section{Introduction}

\IEEEPARstart{M}{ulti}-object tracking (MOT) is a fundamental vision task that underpins a broad range of applications from autonomous vehicle guidance \cite{geiger2012we,du2018unmanned,yu2020unmanned} to smart city management~\cite{milan2016mot16,CIAPARRONE202061}. With RGB cameras as input, typical MOT systems approach the problem by first detecting and localizing objects of interest in each frame, followed by a tracking phase where detected objects from consecutive frames are associated. As a result of this track-by-detection approach, the frame rate of the input, which directly affects the degree of change to an object perceived due to relative motion and deformation, can have a significant effect on the efficacy of an MOT algorithm, especially under high-speed scenarios.

\begin{figure}[tbp]
    \includegraphics[width=\linewidth]{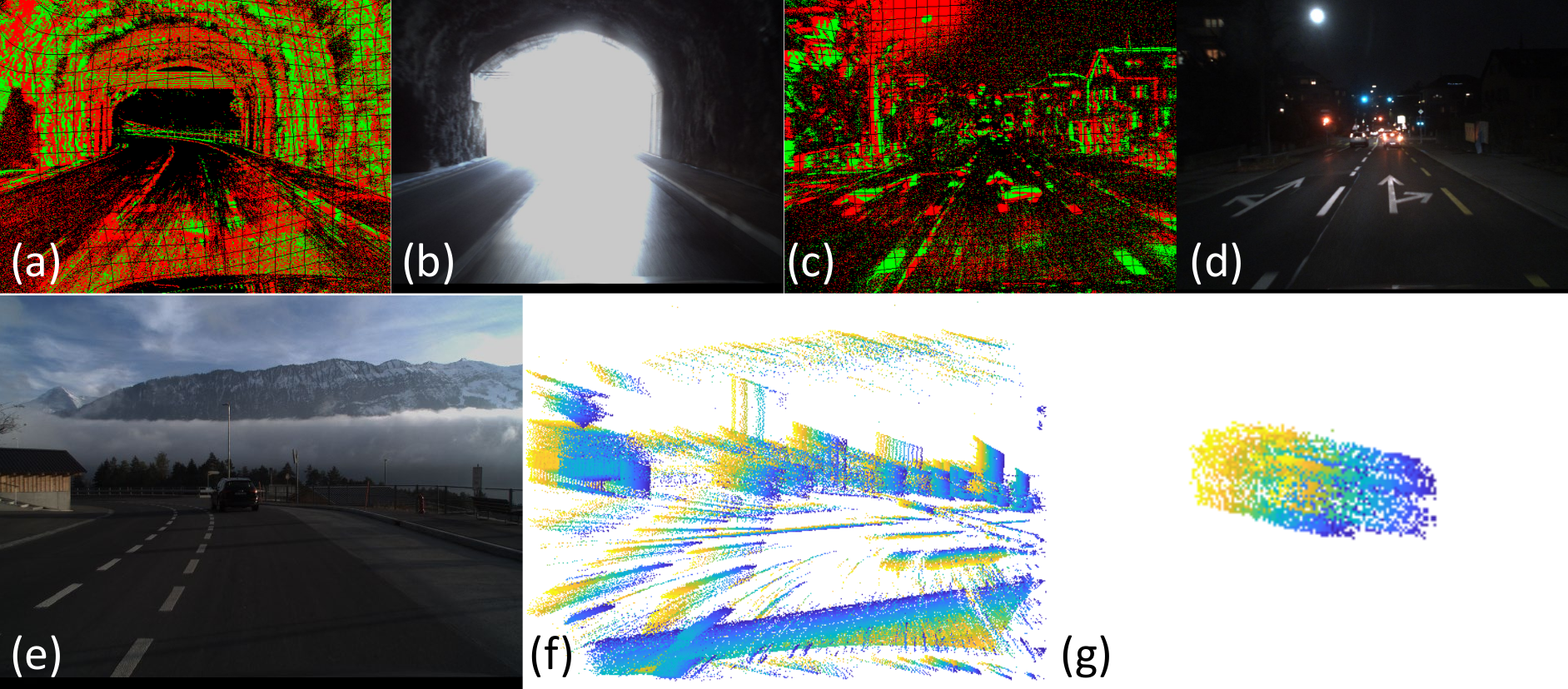}
    \centering
    \caption{Event camera output. (a)-(d) Objects are captured under high contrast and low light conditions. (e)-(f) The same scene captured by an RGB camera and a moving event camera. (g) Events surrounding the car after background events are removed.}
    \label{fig:eventcamera}
\end{figure}

Event cameras, with their asynchronous sensing capabilities, have recently emerged as a promising sensing technology that serves either as a replacement input~\cite{gao2022remot,zhang2022spiking,chen2020end} or as a complementary input to an RGB camera~\cite{gehrig2020eklt,zhang2021object} for object tracking. Instead of measuring the absolute brightness of all pixels synchronously at a fixed frame rate, event cameras respond asynchronously to brightness changes by generating event spikes at the time when and in the pixels where changes occur \cite{lichtsteiner2008128,brandli2014240}. As a result, event cameras naturally produce traces of events that follow the trajectory of an object with high temporal resolution (\cref{fig:eventcamera}g), compared with RGB cameras. Furthermore, in real-world settings with unforgiving lighting conditions, the high dynamic range (in the order of \SI{140}{\decibel}) of event cameras also allows them to operate when typical RGB cameras can easily be handicapped (\cref{fig:eventcamera}a-d).

However, despite the promising technological advantages in theory, simultaneously tracking multiple objects against a complex background using event cameras remains an open challenge in practice. In these complex real-world scenarios, vision events are produced as a result of the relative motions among the objects of interest, the background, and the camera at the same time. A successful MOT algorithm must be able to identify the areas where the events are relevant to the objects of interest and to track the objects in spite of noisy background in their vicinity (\cref{fig:eventcamera}e-f).

In this study, we present \name, an event-based multi-object tracking method that addresses these challenges. By integrating spiking neural networks (SNNs) into a Siamese architecture, \name effectively extracts sparse spatiotemporal features from the binary input derived from the event camera. These feature segments are then used to infer the motion of the tracked objects at high frequency and maintain their identities. In parallel, a YOLO-based detector identifies objects of interest at frame rate across the entire visual field from the camera, and the resulting bounding boxes are associated with the motion tracks from the Siamese network to produce final tracking results.

Furthermore, to promote research in event-based object tracking, we have built DSEC-MOT, the first event-based real-world MOT benchmark. Our dataset builds upon DSEC \cite{gehrig2021dsec}, one of the most challenging event-based datasets to date, containing degraded lighting conditions and heavy traffic. We provided a semi-automated annotation tool and meticulously labeled \num{37080} boxes for \num{7} classes of common traffic entities. Experimental results demonstrate that \name outperforms existing state-of-the-art trackers across a wide range of performance metrics. We also conducted comprehensive analyses to examine the influence of backbone network variations, detector efficacy, and parameter changes on tracking accuracy, thereby evaluating the effectiveness of \name. As such, we consider the main contributions of this work are in the following areas:
\begin{itemize}[leftmargin=*,parsep=0px]
\item We propose a novel MOT network architecture that incorporates the SNNs with SRM neurons into Siamese architecture to extract sparse spatiotemporal features for tracking multiple objects against a complex background.
\item We present a large-scale event-based MOT benchmark based on DSEC for a fair evaluation of trackers. Our annotations cover real-world challenging scenarios where tracking suffers from heavy occlusions, extensive trajectory intersections, uniform appearance, and objects out of view.
\item Our SpikeMOT outperforms state-of-the-art trackers throughout most metrics under real-world scenarios.
\end{itemize}

\section{Related Work}

\subsection{Multi-Object Tracking}
MOT has been dominated by the track-by-detection paradigm \cite{xu2021segment,yin2020unified,wang2020towards,xu2019spatial}, where object instances are first detected and then associated over time based on their spatiotemporal coherence and visual coherence. The trackers based on the spatiotemporal proximity \cite{bewley2016simple,han2022mat,bochinski2017high} are fast because of their simplicity, but they struggle to handle crowded scenes effectively. Conversely, trackers that rely on appearance proximity \cite{fischer2023qdtrack,bergmann2019tracking,zhang2021fairmot,leal2016learning} excel in long-range association and re-identification tasks by leveraging instance appearance similarity. However, these trackers face challenges when objects share similar appearances \cite{sun2022dancetrack}. DeepSORT \cite{wojke2017simple} uses both appearance cues and motion cues for object associations and thus objects can be linked correctly when they are occluded and reappear later. Inspired by DeepSORT, our tracker learns both motion coherence and appearance coherence to update tracklets, which helps reduce the occurrence of identity switches and improve association accuracy.

Siamese networks have emerged as a popular choice for object tracking, owing to their ability to achieve a balance between accuracy and efficiency \cite{cao2023towards,javed2022visual,chen2022siamban,dong2022adaptive,li2018high}. Our tracker is also based on the Siamese network. Siamese trackers determine the similarity between instance templates and search regions by using a similarity function that is learned offline~\cite{hu2023siammask,chen2022high,shuai2021siammot,zhang2019deeper,zhu2018distractor}. The pioneering work of this similarity learning is SiamFC \cite{bertinetto2016fully} which is followed by many studies aimed at improving its accuracy and efficiency \cite{guo2020siamcar,voigtlaender2020siam,li2019siamrpn++,fan2019siamese}. 

However, their performance in low-light conditions and rapid motion situations remains unsatisfactory due to the limitations of RGB cameras. Thus, there is a growing interest in event cameras. For example, \cite{gehrig2020eklt} extracted features from frames and tracked them using events. \cite{zhang2021object} fused features from both frames and events to track a single object. \cite{chen2020end} used an adaptive time surface formulation to encode events into frames, and demonstrated multi-object tracking. Apart from these methods, multiple works tackle the object tracking problem using event cameras~\cite{mitrokhin2018event,li2019robust,ramesh2020tld,barranco2018real,chen2020end,gao2022remot}. But existing event-based trackers are still confined to controlled scenarios where object trajectories rarely intersect, the number of objects is limited, the background is absent, object contours are distinct and undeformable, scene durations are brief, and re-identification is not necessary. Additionally, the current evaluation metrics are inadequate, emphasizing detection while overlooking association.

Our tracking method is the first to undergo rigorous validation in challenging real-world scenarios involving severe occlusions, frequent trajectory intersections, and the need for long-term object re-identification. Through a comprehensive set of evaluation metrics covering both detection and association accuracies, our approach achieves state-of-the-art tracking performance.

\subsection{Event-based MOT Dataset}
Image-based MOT datasets, such as Waymo \cite{sun2020scalability}, KITTI \cite{geiger2013vision}, nuScenes \cite{caesar2020nuscenes}, and MOT Challenge \cite{milan2016mot16,dendorfer2020mot20}, are adopted widely to evaluate image-based trackers. However, event-based datasets are considerably less abundant in comparison. As event-based trackers are increasingly adopting machine learning techniques, large-scale event-based datasets are emerging. For example, Automotive Detection Dataset \cite{de2020large} contains 39 hours of driving scenes including urban, highway, suburbs, and countryside environments, which was later expanded to 1MP-ADD \cite{perot2020learning} for the object detection task. DDD17 \cite{binas2017ddd17} and DDD20 \cite{hu2020ddd20} have total 51 hours of driving scenes for the task of steering angle prediction. MVSEC \cite{zhu2018multivehicle}, recorded by two event cameras from various platforms such as a handheld rig, flying hexacopter, car, and motorcycle, serves the purpose of optical flow estimation. FE108 \cite{zhang2021object}, designed for single object tracking, contains 108 sequences and includes diverse lighting conditions and rapid object motion situations. An enhanced version of this dataset, named FE240hz, extends its scope by featuring more complex and challenging scenarios, such as extreme camera motion and strobe light effects. VisEvent \cite{wang2021visevent} contains 820 short object sequences with camera motion and cluttered background for single object tracking. Despite the large-scale and real-world nature of these emerging event-based datasets, none of them are targeted at MOT tasks and all lack the necessary MOT groundtruth, i.e., tracklet IDs. 

DSEC \cite{gehrig2021dsec} was recorded by high-resolution cameras (two 640$\times$480-pixel event cameras and two 1440$\times$1080-pixel RGB cameras) under a variety of lighting conditions, such as night, sunrise, and sunset. However, this dataset lacks groundtruth bounding boxes. To address this limitation, we developed a semi-automated annotation tool and built the DSEC-MOT benchmark, the first real-world event-based MOT benchmark containing severe occlusions, extensive trajectory intersections, and long-term object re-identification, to offer a fair evaluation for event-based MOT trackers.

\subsection{Spiking Neural Networks for Visual Tasks}
SNNs are naturally suitable for addressing event-driven vision challenges due to their inherent asynchronous computational abilities and adeptness in processing sparse spatiotemporal data~\cite{schuman2022opportunities}. Inspired by biological neural networks, SNNs update each neuron's membrane potential based on its previous membrane potential and incoming event signals. These signals are weighted by the synapses that connect the neurons. These characteristics allow SNNs to incorporate both spatial and temporal details into their framework. Unlike conventional artificial neural networks (ANNs) that depend on continuous activations, SNNs rely on discrete spike events for inter-neuronal communication. Owing to the event-driven paradigm and the spatiotemporal dynamics, SNNs offer a promising method for complex motion-related visual tasks, such as object tracking \cite{luo2021siamsnn,zhang2022spiking}, motion segmentation \cite{parameshwara2021spikems}, gesture recognition \cite{amir2017low}, and optical flow \cite{paredes2019unsupervised,lee2020spike}. However, the non-differentiable nature of spiking functions hampers the trainability of SNNs, resulting in their accuracy trailing behind that of current ANNs. Therefore, the approach employing surrogate gradients is explored to allow direct training of SNNs by backpropagation \cite{neftci2019surrogate, fang2021incorporating, wu2018spatio}, capturing the spatiotemporal dynamics of SNN systems. Recent research has achieved accuracy levels competitive with ANN models \cite{yin2021accurate, yao2023attention}. For example, \cite{fang2021incorporating} reported 99.72\% accuracy on MNIST dataset, comparable to SOTA ANN accuracy \cite{mazzia2021efficient}. Therefore, we use backpropagation with surrogate gradients to train our SNN.

A variety of neuron models have been proposed in the field of SNNs, including the Hodgkin-Huxley (HH) model \cite{hodgkin1952quantitative}, the leaky integrate-and-fire (LIF) model \cite{gerstner2014neuronal}, and the Spike Response Model (SRM) \cite{gerstner2001framework, gerstner2002spiking}. While the HH model pioneered in its depiction of neuronal dynamics through differential equations, it proves too complex for practical applications. Contrarily, the LIF model, characterized by its simplicity and computational efficacy, is the most widely used spiking neuron model. However, it falls short in simulating the full spectrum of realistic spiking neuron traits \cite{izhikevich2004model}. In this context, the SRM offers a simple yet versatile neuronal archetype, adept at offering nuanced portrayals of neuron dynamics \cite{gerstner2014neuronal, masquelier2016microsaccades, gerstner2002spiking}. The SRM neuron, functioning as a derivative approximation of the LIF model, describes the membrane potential by kernels' integration over incoming and outward spikes. The membrane potential ascent culminates in the discharge of a spike when reaching a threshold. Following this discharge, a refractory signal briefly inhibits the neuron's activity. This soft reset mechanism in SRM neurons differs from the hard reset in LIF neurons, where the potential immediately returns to a baseline resting level. Such inhibitory dynamics endow SRM neurons to memorize longer-term dependencies and yield a diverse array of spiking patterns in comparison with LIF neurons, resulting in enhanced accuracy \cite{yin2023accurate}. The SNN \cite{shrestha2018slayer} employing SRM neurons achieved 99.20\% accuracy on the N-MNIST dataset, outperforming the SNN utilizing LIF neurons, RNN, and LSTM models, all of which possess significantly more parameters \cite{he2020comparing}. In the pursuit of equilibrium between the computational efficiency and the spectrum of output spikes, we employ the SRM neuron model in our SNN.

The closest work related to ours is STNet \cite{zhang2022spiking} that designed a spiking network integrated with a transformer, where the transformer provides global spatial information and an SNN extracts temporal cues. Notably, their work primarily centered around the object detection task. In light of the available literature, our work stands as the pioneering effort in achieving a competitive performance in the field of event-based MOT, particularly under the formidable real-world scenarios.

\section{\name}
\begin{figure}[tbp]
    \includegraphics[width=\linewidth]{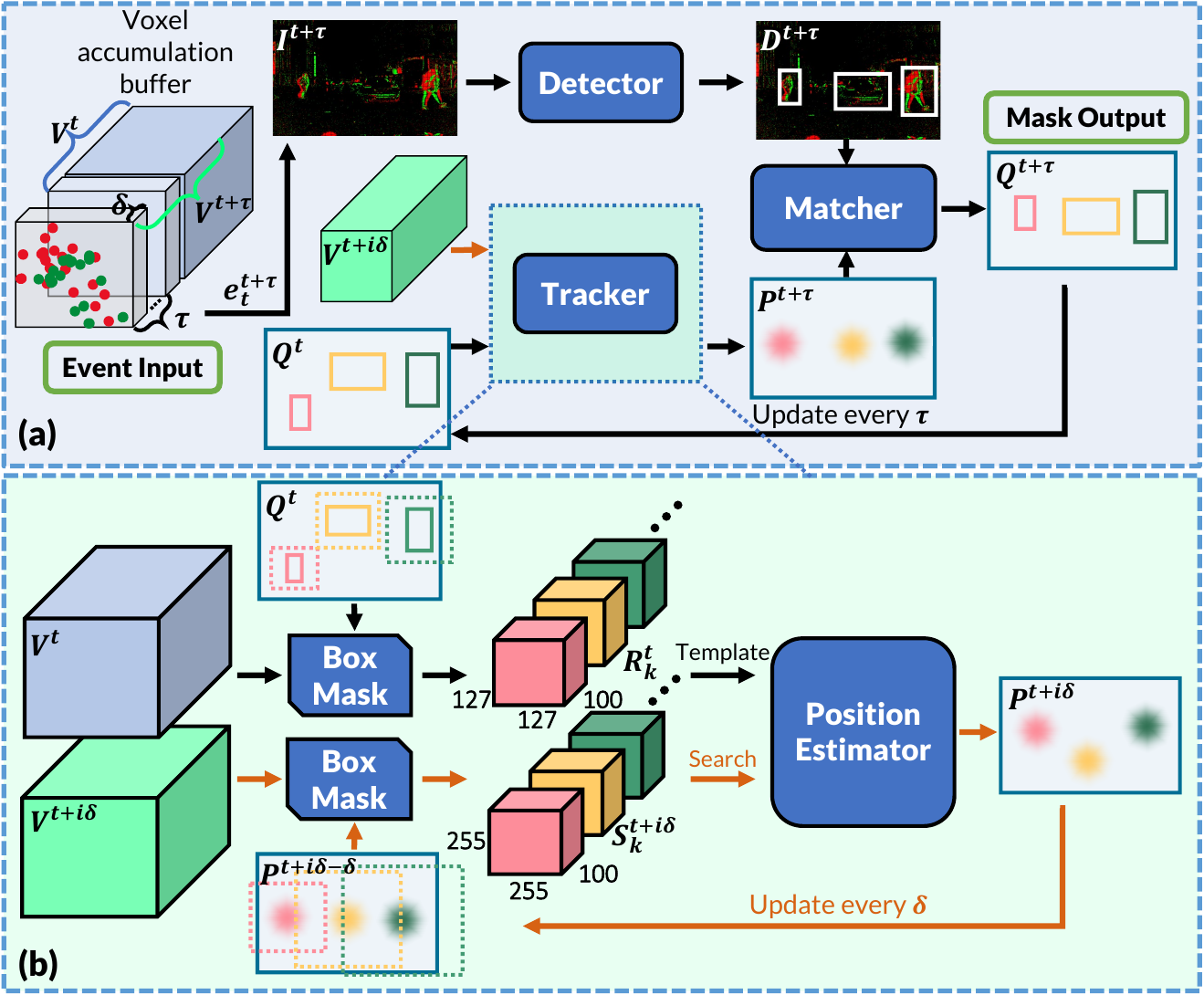}
    \centering
    \caption{Overview of \name. (a) Tracklets are updated every $\tau$ period by associating the latest tracked coordinates with the detected objects; (b) Tracker updates object coordinates every $\delta$ period leveraging the high temporal resolution of event cameras.}
    \label{fig:spikemot}
\end{figure}

The \name comprises three elements: (i) an object tracker, (ii) an object detector, and (iii) a tracklet matcher (\cref{fig:spikemot}a). While the detector identifies objects of interest and updates object templates at a lower frequency ($1/\tau$) comparable to an RGB camera frame rate, the tracker maintains continuity of object identities across frame duration by updating the coordinates of the tracked objects at a higher frequency ($1/\delta$). The matcher associates the detector's objects with the tracker's objects to update existing tracklets, initialize new tracklets, and eliminate tracklets of disappeared objects.

\subsection{Overview of Object Tracking}
\cref{fig:spikemot}a shows an overview of \name's operation from time $t$ to $t+\tau$, which corresponds to the duration of $1$ frame in an equivalent RGB camera. Let $H, W$ be the spatial resolution of the event camera and $e_{t_1}^{t_2}$ be a $2 \times H \times W \times (t_2-t_1)$ binary sparse tensor formed by setting $1$ for the events $(p, y, x, t)$ produced from time $t_1$ to $t_2$. Define an \emph{event voxel} $V^{t}$ as a dense tensor converted from the sparse tensor $e_{t-\zeta}^{t}$, where $\zeta$ is a duration parameter of the voxel $V$. $V$'s temporal axis is quantized into bins of a duration of $\delta$ (where $\delta$ represents voxel granularity). Then, at every $i$-th period of $\delta$, where $i=0,1,\ldots,\tau/\delta-1$, the tracker updates the current estimated positions $P^{t+i\delta}$ of the tracked instances by using the tracked instances $Q^t$ and the updated event voxel $V^{t+i\delta}$ at that time. In parallel, the detector uses $e_t^{t+\tau}$ to produce a set of detected objects of interest $D^{t+\tau}$ for this frame. Towards the end of the frame, the latest positions $P^{t+\tau}$ of the tracked instances are matched with the detected objects $D^{t+\tau}$ to update the tracked instances $Q^{t+\tau}$, which will be used for tracking in the next frame. Formally,
\begin{equation}
    \label{eq1}
    \begin{aligned}
        P^{t+\tau} &= f_{T}(Q^t, V^t, e_t^{t+\tau}; \theta_{T}) \\
        D^{t+\tau} &= f_{D}(e_t^{t+\tau}; \theta_{D}) \\
        Q^{t+\tau} &= f_{M}(P^{t+\tau}, D^{t+\tau})
    \end{aligned}
\end{equation}
where $f_{T}$ is the Siamese position estimator with the parameters $\theta_{T}$, $f_{D}$ is the detector with the parameters $\theta_{D}$, and $f_{M}$ is the matcher function.

\cref{fig:spikemot}b shows the operation of SpikeMOT tracker during a period of $\delta$, which is centered around a Siamese network that performs position estimation for each tracked instance. At time $t$, the instance template voxels $\{R_k^t\}_{k=1}^{N}$ are prepared by the box mask that crops the event voxel $V^t$ according to the tracked instances $Q^t$ and reshapes the cropped sub-voxels into a uniform size. At the time $t+i\delta$ when the event voxel $V^{t+i\delta}$ has been updated since the time $t+i\delta-\delta$, the contextual search voxels $\{S_k^{t+i\delta}\}_{k=1}^{N}$ are prepared by the box mask that crops the current event voxel $V^{t+i\delta}$ with expanded regions around the estimated positions $P^{t+i\delta-\delta}$ obtained from the previous iteration and reshapes the cropped sub-voxels into a uniform size. Given an instance template voxel $R_k^t$ at time $t$ and a contextual search voxel $S_k^{t+i\delta}$ at time $t+i\delta$, the Siamese position estimator searches for the particular instance around its previous location and obtains the updated positions $P^{t+i\delta}$ at present time.

\subsection{Position Estimator}
\begin{figure*}[tbp]
    \includegraphics[width=\linewidth]{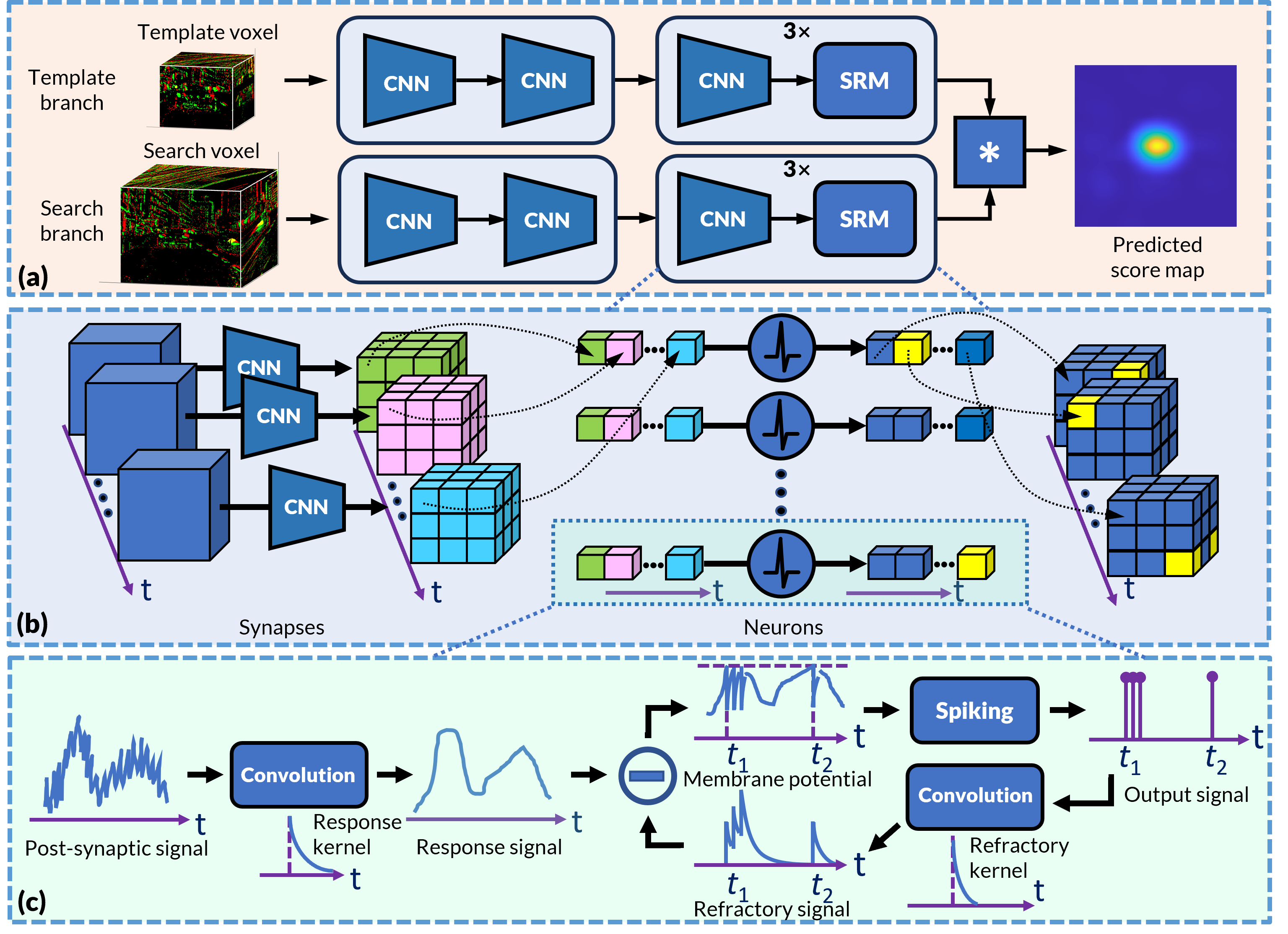}
    \centering
    \caption{Siamese position estimator. (a) the architecture of the proposed Siamese network, with the template and the search branches incorporating two-stage SNN models. (b) The fundamental unit of the SNN model, encompassing the synapses represented by CNNs and the neurons modelled by SRMs. (c) Spiking neuron model, wherein the ascent of internal membrane potential culminates in the discharge of a spike once a threshold is reached. The emitted spike is then convolved by the refractory kernel to update the refractory signal, which in turn suppresses the subsequent response signal.}
    \label{fig:siam}
\end{figure*}

\cref{fig:siam}a illustrates the architecture of the proposed Siamese network that runs at the heart of the \name. The proposed Siamese network encompasses the template branch and the search branch, each incorporating our SNN model organized into two stages. The initial stage consists of two CNN blocks designed for feature dimension reduction. Meanwhile, the subsequent stage comprises three sequential blocks responsible for the extraction of spatiotemporal features. Leveraging the SNN's inherent spatiotemporal reasoning capabilities, the template and search branches of the Siamese network transform motion voxels into sparse spatiotemporal features, facilitating the final feature association procedure and yielding predictions for target positions.

\subsubsection{SNN Model}
\cref{fig:siam}b shows one block in the second stage of the SNN model, which serves as the fundamental unit for the overall SNN model. Each block contains the synapse network (CNNs) responsible for extracting spatial features from input sequences, followed by the neurons (SRMs) responsible for abstracting temporal features at each pixel. By adopting this structure, the SNN model thus leverages the combined benefits of the accuracy embedded in CNN architectures and the spatiotemporal reasoning capabilities inherent to SNNs. Different from the previous SRM-based SNN model in \cite{gerstner1995time}, our model eliminates pre-synaptic kernels located preceding the synapse network and incorporates post-synaptic kernels within the neurons. This modification significantly reduces computational costs. The detailed architecture of this SNN model is summarized in \cref{tab:archi}.

\begin{table}[!h]
  \caption{Architecture of the SNN Model.}
  \centering
  \begin{tabular}{@{} c c c c c c @{}}
    \hline
     Layer            & Kernel         & Stride       & Channels & \splittab{Template}{feature}  & \splittab{Search}{feature} \\ 
    \hline
                    &                  &              &   2 & 127 $\times$ 127 & 255 $\times$ 255 \\
     Conv           &   11 $\times$ 11 & 2 $\times$ 2 &  96 &  59 $\times$  59 & 123 $\times$ 123 \\
     MaxPool        &    3 $\times$  3 & 2 $\times$ 2 &  96 &  29 $\times$  29 &  61 $\times$  61 \\
     \hline
     Conv           &    5 $\times$  5 & 1 $\times$ 1 & 256 &  25 $\times$  25 &  57 $\times$  57 \\
     MaxPool        &    3 $\times$  3 & 2 $\times$ 2 & 256 &  12 $\times$  12 &  28 $\times$  28 \\
     \hline
     Conv           &    3 $\times$  3 & 1 $\times$ 1 & 384 &  10 $\times$  10 &  26 $\times$  26 \\
     SRM            &    -             & -            & 384 &  10 $\times$  10 &  26 $\times$  26 \\
     \hline
     Conv           &    3 $\times$  3 & 1 $\times$ 1 & 384 &   8 $\times$   8 &  24 $\times$  24 \\
     SRM            &    -             & -            & 384 &   8 $\times$   8 &  24 $\times$  24 \\
     \hline
     Conv           &    3 $\times$  3 & 1 $\times$ 1 & 256 &   6 $\times$   6 &  22 $\times$  22 \\
     SRM            &    -             & -            & 256 &   6 $\times$   6 &  22 $\times$  22 \\
    \hline
  \end{tabular}
  \label{tab:archi}
\end{table}

\subsubsection{Spiking Neuron Model}
We choose to utilize the SRM as our spiking neuron model, owing to its simplicity, versatility, and long-term dependency compared to alternative models. Our SRM neuron accumulates post-synaptic signals from CNNs over time, resulting in a continuous-time and smooth membrane potential that is then encoded into sparse spikes. As illustrated in \cref{fig:siam}c, the post-synaptic signal within the $l_{th}$ layer, denoted as $o^{l}(t)$, undergoes filtering into response signals through a convolution operation with a response kernel denoted as $\varepsilon$. Similarly, the output signal for the next $(l+1)_{th}$ layer, expressed as $s^{l+1}(t)$, is subjected to an analogous convolution process employing a refractory kernel denoted as $\upsilon$, yielding refractory signals. The response and refractory kernels are in the form of the simplest first-order low-pass filter to minimize the computational load. Subsequently, the response signals are inhibited by the refractory signals, resulting in the membrane potential. If the membrane potential surpasses the threshold $V_{th}$, a spike signal is generated. This process is formally described as follows.
\begin{equation}
    \label{eq6}
    \begin{gathered}
        u^{l}(t)=\int_{0}^{\infty}\varepsilon(t-\tau)o^{l}(\tau)d\tau - \int_{0}^{\infty}\upsilon(t-\tau)s^{l+1}(\tau)d\tau \\
        \varepsilon=\frac{1}{\tau_s}e^{-t/\tau_s}H(t) \\
        \upsilon=\frac{1}{\tau_r}e^{-t/\tau_r}H(t)
    \end{gathered}
\end{equation}
\begin{equation}
    \label{eq7}
    \begin{aligned}
        s^{l+1}(t)       &= f_s\Big(u^l(t)\Big) \\
        f_{s}(u): s(t_i) &=
            \left\{
            \begin{aligned}
                \delta(t - t_i), \quad & if \ u(t_i) \geq V_{th} \\
                0, \quad               & otherwise
            \end{aligned}
            \right.
    \end{aligned}
\end{equation}
where $H(t)$ represents the Heaviside step function, $\tau_s$ and $\tau_r$ denote neuron time constants for response and refractory signals, respectively, $u^l(t)$ is the membrane potential, $f_s$ is a threshold function with the threshold of $V_{th}$, and $\delta$ is the Dirac delta function.

\subsection{Detection \& Matching}
\name's detector leverages the one-stage YOLOv5~\cite{glenn_jocher_2020_4154370} for object detection tasks. Compared to two-stage detection models, such as Faster R-CNN \cite{ren2015faster}, the one-stage YOLOv5 demonstrates superior efficiency and effectiveness in the detection of small objects. The detector was trained utilizing a COCO pre-trained model on event frames, where objects can be sensed amidst degraded scenes (\cref{fig:eventcamera}a, c). Notably, the detector tailored to event frames exhibits commendable performance in testing scenarios, demonstrated in \cref{tab:yolo_performance}. These metrics collectively affirm the adeptness of YOLOv5 in discerning objects within event frames. The event frames accumulated from the events in every frame time have the same structural format as conventional RGB images. Within this format, when a positive event is registered, the red channel of the event frame corresponding to the event's spatial location is assigned $1$. Similarly, when a negative event arrives, the green channel of the event frame is set to $1$.

\begin{table}[!h]
  \caption{YOLOv5's Performance on DSEC-MOT Test Set.}
  \centering
  \begin{tabular}{@{} c c c c @{}}
    \hline
     Precision & Recall & mAP@0.5 & mAP@(0.5:0.95) \\ 
    \hline
          0.87 & 0.69   & 0.78    &   0.54 \\
    \hline
  \end{tabular}
  \label{tab:yolo_performance}
\end{table}

In the final stage of \name, the tracklet matcher is responsible for ascertaining the latest coordinates, size, and tracklet ID of target entities by associating the entities derived from the detector with those updated by the tracker, similar to that used in~\cite{bewley2016simple,wojke2017simple,shuai2021siammot}. Firstly, the Hungarian algorithm is employed to correlate detected objects with their corresponding tracked counterparts, utilizing the Hausdorff distance \cite{huttenlocher1993comparing} as the basis for this association. Subsequently, the formed associations undergo validation or dismissal contingent on the outcomes generated by the Siamese position estimator. In comparison with solely relying on the Siamese position estimator for the association, the utilization of the Hungarian algorithm reduces the dependence on the Siamese position estimator and significantly alleviates the computational load. Following the validation step, a solver component executes updates to the tracklets that have successfully matched with the detected entities. Additionally, it generates new tracklets when there exists an unpaired detection extending over a consecutive span of frames and eliminates those tracklets that remain unmatched over a continuous frame duration.

\subsection{Training Samples}
To enhance the tracker's discriminative capacity, three categories of antagonistic voxel pairs involving template-search voxels have been incorporated into the training dataset. These negative pair voxels derive from three distinct origins: 1) voxels corresponding to objects belonging to the same category but distinct objects; 2) voxels emanating from different categories; or 3) voxels emerging from an individual object and its background. These three categories constitute 25\%, 12.5\%, and 12.5\% of the dataset, correspondingly, while the residual voxels form positive pairs sourced from the same object.

The inclusion of negative pair voxels arising from distinct categories serves to train the tracker the ability to circumvent unintended drift towards arbitrary surrounding objects when the target becomes completely occluded. Negative pair voxels originating from objects within the same category but different objects train the tracker the capacity to concentrate on fine-grained representations and to discriminate against semantic distractors based on their motion characteristics.

\subsection{Loss} 
The Siamese position estimator's loss function is described as the weighted sum of individual loss $l_i$, corresponding to each position $u_i$ within the predicted score map, as elucidated by Eqs. \ref{eq2}-\ref{eq5}. The groundtruth label $y_i$ at each position $u_i$ is defined by Eq. \ref{eq2}, where the labels are $1$ if their positions fall within the radius $R$ of the target's groundtruth center $c_y$. Conversely, for the negative pairs of template-search voxels where the target is absent, all groundtruth labels $y_i$ are uniformly set to $0$.
\begin{equation}
    y_i =
    \left\{
    \begin{aligned}
        1, \quad & \|u_i-c_y\| \leq R \\
        0, \quad & \|u_i-c_y\| > R
    \end{aligned}
    \right.
    \label{eq2}
\end{equation}

We apply a weight $w_i$ on individual loss $l_i$, which is defined by Eq. \ref{eq3}. It is noteworthy that the sum of the weights inside the radius $R$ equals to the sum outside, thereby eliminating the imbalance of the losses between regions inside and outside the groundtruth radius.
\begin{equation}
    w_i =
    \left\{
    \begin{aligned}
        \frac{0.5}{\sum[y_i=1]}, \quad & \|u_i-c_y\| \leq R \\
        \frac{0.5}{\sum[y_i=0]}, \quad & \|u_i-c_y\| > R
    \end{aligned}
    \right.
    \label{eq3}
\end{equation}

The loss computation of negative pairs of template-search voxels is described by Eq. \ref{eq4}, where $v_i$ is the predicted score at the position $u_i$ and $\sigma$ is the sigmoid function.

\begin{equation}
    \label{eq4}
    L_{neg} =\sum\limits -w_i\log(1-\sigma(v_i))
\end{equation}

The loss computation of positive pairs of template-search voxels is described by Eq. \ref{eq5}, where $c_v$ is the peak position of the predicted score map and $\epsilon$ is an infinitesimal value.

\begin{equation}
    \label{eq5}
    \begin{aligned}
            & \|u_i-c_v\| > R: \\
            & l_i = -y_i\log(\epsilon) \\    
            & \|u_i-c_v\| \leq R: \\
            & l_i = -[y_i\log(\sigma(v_i))+(1-y_i)\log(1-\sigma(v_i))] \\
            & L_{pos} = \sum\limits w_i l_i
    \end{aligned}
\end{equation}

\section{DSEC-MOT}
\begin{figure*}[!h]
    \includegraphics[width=\linewidth]{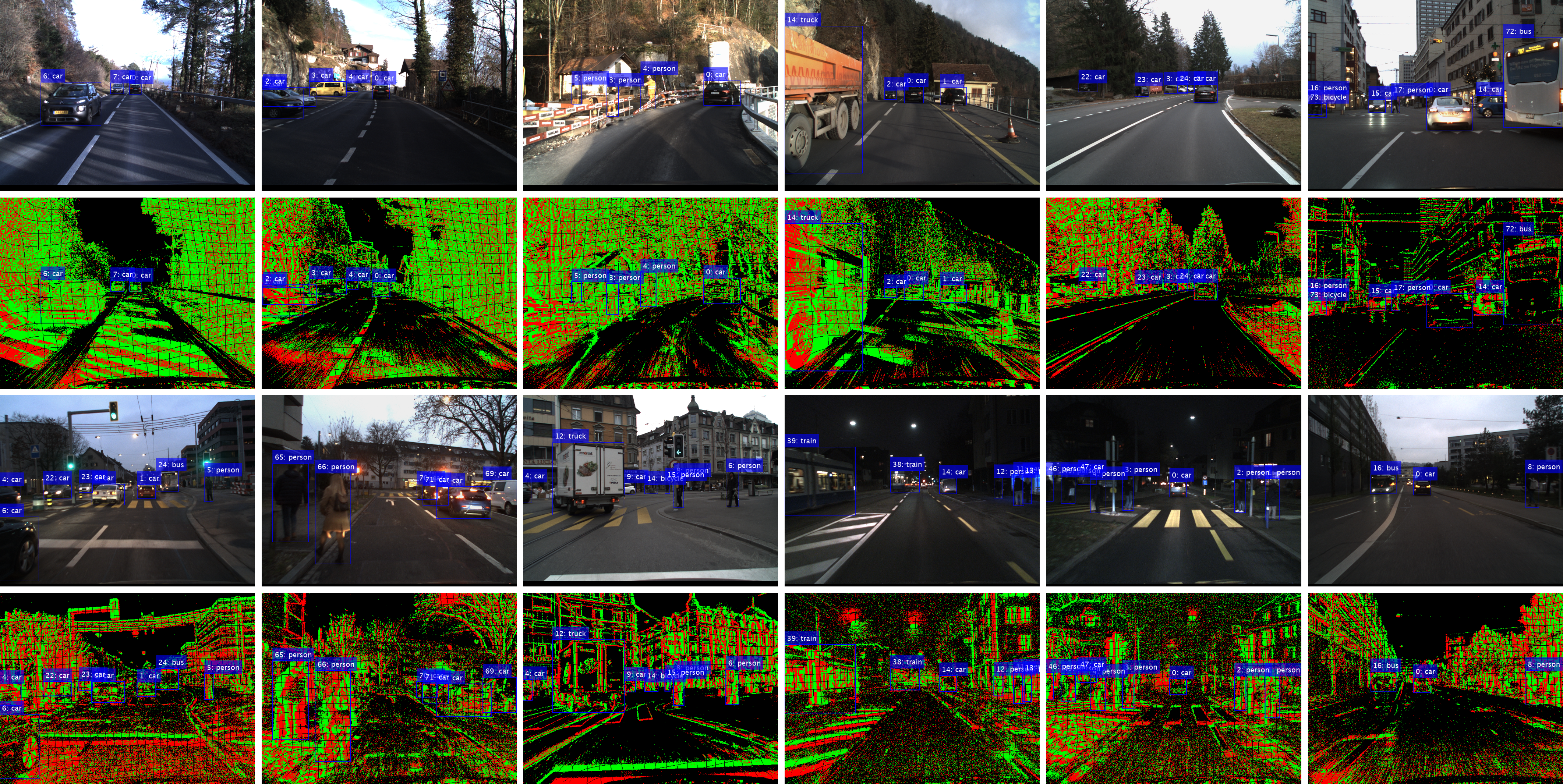}
    \centering
    \caption{An overview of the scenes in each sequence of DSEC-MOT, which include cluttered background and heavy occlusion. The 1st and the 3rd rows show the RGB images of each sequence, while the 2nd and the 4th rows demonstrate the corresponding event-based representations.}
    \label{fig:dataoverview}
\end{figure*}

The DSEC-MOT dataset has been established as a benchmark for the event-based MOT tasks. The purpose of launching this dataset is to provide an objective measure of event-based MOT trackers. A visual representation of the DSEC-MOT is shown in \cref{fig:dataoverview}, with a detailed description provided in \cref{tab:dataset}.

\begin{table*}[!h]
  \centering
  \caption{An Overview of the Annotations in Each Sequence of DSEC-MOT.}
  \begin{tabular}{@{} c c c c | c c c c c c c @{}}
    \hline
     Seq. & Len. (sec.) & Tracks & Boxes & Car & Person & Bicycle & Motorcycle & Bus & Truck & Train \\ 
    \hline
     interlaken\_00\_a            &   57.1  &   9  &   1,387 &   1,342  &     0  &      0  &      0  &    0  &   45  &     0 \\
     interlaken\_00\_b            &   80.8  &  16  &   1,601 &   1,571  &    30  &      0  &      0  &    0  &    0  &     0 \\
     interlaken\_00\_c            &   26.8  &   9  &     928 &     750  &   117  &      0  &      0  &    0  &   61  &     0 \\
     interlaken\_00\_e            &   99.5  &  48  &   3,293 &   3,119  &    94  &     11  &      0  &   69  &    0  &     0 \\
     zurich\_city\_01\_d          &   39.7  &  63  &   3,612 &   3,149  &   172  &      9  &     29  &  197  &   56  &     0 \\
     zurich\_city\_01\_e          &   99.5  &  93  &   6,204 &   4,410  & 1,479  &    196  &     15  &   57  &   47  &     0 \\
     zurich\_city\_04\_b          &   13.4  &  19  &   1,123 &     534  &   231  &     58  &     32  &    0  &  268  &     0 \\
     zurich\_city\_09\_c          &   66.1  &  46  &   3,667 &   1,703  &   737  &      0  &     27  &    0  &    0  & 1,200 \\
     zurich\_city\_09\_d          &   84.9  &  60  &   4,886 &   4,026  &   595  &    265  &      0  &    0  &    0  &     0 \\
     zurich\_city\_14\_b          &   28.8  &  20  &   1,181 &     789  &   313  &      0  &      0  &   79  &    0  &     0 \\
    \hline
    Total training                &  596.6  & 383  &  27,882 &  21,393	& 3,768	 &    539  &    103	 &  402	 &  477	 & 1,200 \\
    \hline
    interlaken\_00\_d             &   99.5  &  23  &   2,687 &   2,441  &   160  &      0  &     21  &    0  &   65  &     0 \\
    zurich\_city\_00\_b           &   73.1  &  96  &   6,511 &   2,223  & 3,563  &    206  &     46  &  441  &    0  &    32 \\
    \hline
    Total testing                 &  172.6  & 119  &   9,198 &   4,664	& 3,723	 &    206  &     67	 &  441	 &   65	 &    32 \\
    \hline
    Total                         &  769.2  & 502  &  37,080 &  26,057  &  7,491 &    745  &    170  &  843  &  542  & 1,232 \\
    \hline
  \end{tabular}
  \label{tab:dataset}
\end{table*}

\subsection{Dataset Statistics}
DSEC-MOT encompasses the tracking of diverse object classes, including cars, persons, bicycles, motorcycles, buses, trucks, and trains. These objects are tracked across a wide spectrum of urban, suburban, and rural environments, covering various lighting conditions such as daylight, dusk, and nighttime scenarios. Notably, the dataset is characterized by sequences incorporating a substantial volume of traffic instances. The dataset is meticulously annotated, encompassing a training set of \num{10} sequences comprising \num{27882} detections and \num{383} distinct object tracks. Additionally, it includes a testing set of \num{2} sequences, comprising \num{9198} detections and \num{119} individual object tracks. It is worth emphasizing that DSEC-MOT is unique in its inclusion of highly congested street views, distinguishing it from existing event datasets \cite{mueggler2017event,zhang2021object}. Consequently, we anticipate that this benchmark will present a formidable and rewarding challenge to the event-based object-tracking community.

\subsection{Labeling Protocol}
Open-source annotation tools like CVAT and LabelImg are widely used for object detection. However, accurately associating detections into coherent tracklets can be challenging when employing these tools. In order to facilitate the manual annotation process, we present a semi-automated approach that preprocesses the raw dataset and identifies potential objects that pertain to a specific tracklet before the subsequent labeling phase.

The proposed approach begins with the calibration of images and events within the DSEC dataset. Notably, the images are captured at a resolution of 1.5 megapixels, whereas the corresponding events are recorded at a resolution of 0.3 megapixels. To ensure congruity, a rectification process is executed to align the pixel arrays of the images and events, allowing the annotations generated on the images to serve as groundtruth for the events. Subsequently, object detection algorithms are applied to the rectified images, initiating the tracklet association process. 

During the process of annotating a tracklet, the linkage of detected bounding boxes across consecutive frames is achieved by means of minimizing the Hausdorff distance, thereby establishing a preliminary association. Subsequently, the bounding box in the current frame corresponding to the identified tracklet is presented to a human annotator for validation. The annotator's task encompasses confirming the association, associating it with another detected bounding box, manually delineating a bounding box in the present frame, or disregarding the proposed association. Following the annotation of a tracklet, a refinement phase is undertaken to manually adjust the bounding boxes, optimizing them to encompass the entirety of object pixels while simultaneously maintaining a compact form.

To ensure the rigorous standards of annotation quality, a quality control process is implemented. All annotations undergo meticulous scrutiny by a separate annotator, and identified errors are rectified through a re-annotation process. The objects occupying an area less than 0.2\% of total pixels (640$\times$480) are excluded from annotation, due to the consideration of image resolution and quality. In cases of partial occlusion, only the visible portions of objects are annotated, while objects that are occluded by at least 90\% or disappear by more than 90\% in the event domain are exempted from annotation, according to their attributes under such occluded conditions. Nevertheless, the identifiers associated with occluded or disappeared objects are meticulously retained to facilitate subsequent re-identification when these objects re-emerge in subsequent frames. This practice ensures the continuity of object tracking and re-establishment of object identities.

\section{Experiments}

In this section, we provide an in-depth analysis of the tracking outcomes achieved by \name on DSEC-MOT and FE240hz \cite{zhang2021object} datasets. These results underscore the remarkable performance of \name in challenging scenarios involving severe occlusions and a significant presence of background events, outperforming contemporary state-of-the-art MOT tracking methods. Additionally, we elucidate the efficacy of our model, attributing its success to the recurrent pathways and the sparse spatiotemporal features derived from the SNN architecture.

\subsection{Datasets}
We compared our proposed \name with state-of-the-art trackers on two datasets, FE240hz and DSEC-MOT. 
The FE240hz is an event-image-based dataset for single object tracking. The resolution of its images and events is 346$\times$260 pixels. It is characterized by intricate and demanding scenarios, such as severe camera motion, low light, and high dynamic range. Additionally, it provides a high annotation frequency of up to $240$ Hz in the event domain. DSEC-MOT is an event-image-based dataset designed for multi-object tracking, with an average of 42 tracks per sequence. The prominent characteristic of this dataset is the severe occlusion, and the challenges revolve around frequent view obstruction occurring between vehicles and vehicles, vehicles and pedestrians, as well as pedestrians and pedestrians. The images and events in DSEC-MOT exhibit a much higher resolution of 640$\times$480 pixels. Besides, all of its sequences are captured by moving cameras, leading to a substantial volume of background events mingled with object events. The analysis of the two event-based datasets is detailed in \cref{tab:data}.

\begin{table*}
  \centering
  \caption{Analysis of the Event-based Datasets Where Trackers Are Compared.}
  \begin{tabular}{@{} c c c c c c c c c @{}}
    \hline
    Dataset & Tracking targets & Class & \splittab{Avg. events}{/ seq.}  & \splittab{Avg. frames}{/ seq.} & \splittab{Avg. length}{/ seq. (sec.)} &  \splittab{Avg. tracks}{/ seq.} & \splittab{Avg. boxes}{/ seq.} &  \splittab{Annotation}{frequency (Hz)}\\
    \hline
    FE240hz  & Daily common objects & 17 &  55.8M  & 1,295 & 68.1  &  1  & 9,998 & up to 240 \\
    DSEC-MOT & Traffic entities      &  7 & 881.8M  & 1,283 & 64.1  & 42  & 3,090 & 20 \\
    \hline
  \end{tabular}
  \label{tab:data}
\end{table*}

\subsection{Evaluation Metrics}
We use the metrics of HOTA \cite{luiten2021hota}, CLEAR \cite{bernardin2008evaluating}, and IDF1 \cite{ristani2016performance} to evaluate \name from different aspects. CLEAR focuses more on detection performance, while IDF1 assesses the ability of identity preservation and focuses more on association performance. Recently HOTA has begun to be used for MOT evaluation. It explicitly balances the assessment of detection and association, and has emerged as the primary metric to evaluate MOT trackers across various well-established benchmarks, including MOT17 \cite{milan2016mot16} and KITTI \cite{geiger2013vision}. HOTA's detection accuracy assesses the alignment between predicted and groundtruth detections, while HOTA's association accuracy measures the ability of a tracker to link detections over time into the same identities. HOTA exponentially averages these two metrics to provide an overall measurement of tracking accuracy. Notably, we are the first to comprehensively evaluate event-based multi-object trackers using the HOTA metric. Therefore, to compare with the previous trackers and follow the current trend, we employ the metrics of HOTA, CLEAR, and IDF1. Previously widely-used statistics are also provided for fine-grained analysis, such as False Positive (FP), False Negative (FN), and ID switch (IDSw).

\subsection{Implementation Details}
\subsubsection{Network} 
The backbone of the Siamese position estimator consists of five blocks, wherein the first two blocks reduce the feature dimensions and the following three blocks extract sparse spatiotemporal features. Each of the first two blocks incorporates a 3D convolutional layer, a 3D batch normalization layer, a GELU activation function, and a 3D max pooling layer. Each of the subsequent two blocks combines a 3D convolutional layer, a 3D batch normalization layer, a GELU activation function, and an SRM layer. The final block integrates a 3D convolutional layer and an SRM layer. \cref{tab:archi} illustrates the parameter dimensions and the feature map specifications of the convolutional and the pooling layers. Additionally, specific parameters of the SRM layer are defined as follows: The time constants of the response kernel $\tau_s$ and the refractory kernel $\tau_r$ are \SI{10}{\milli\second} and \SI{1}{\milli\second}, respectively. The spiking threshold is 1.0. Furthermore, the template and search voxels, which are extracted and rescaled from event voxels, have the dimensions of $[2, 127, 127, 100]$ and $[2, 255, 255, 100]$, respectively.

\subsubsection{Training}
We implemented the proposed tracker in PyTorch. The tracker is trained from scratch using an equal number of positive and negative voxel pairs. We use AdamW as the optimizer with a weight decay of $10^{-4}$ and train the tracker for $50$ epochs with the batch size of $4$. The learning rate is initialized as $10^{-3}$, and subsequently decayed exponentially to $10^{-4}$. We employ an exponentially decaying probability density function to calculate the surrogate gradient of the non-differentiable spiking threshold function $f_s$~\cite{wu2018spatio}. The derivative of the convolution operation within the SRM neuron is implemented by a correlation operation~\cite{shrestha2018slayer}. The detector is trained using a COCO pre-trained model for $100$ epochs with the batch size of $16$. 

\subsubsection{Inference}
Tracklets are established by the detector upon observing three consecutive detections and subsequently propagated by the tracker. Tracklets undergo removal if they remain unassociated with any detections for a continuous span of $100$ frames. The tracker executes periodic updates on each tracklet at \SI{10}{\milli\second} intervals, while the detector offers detections at \SI{50}{\milli\second} intervals. Specifically, both the template and search voxels comprise $100$ slices, which represent the voxel duration. Each slice accumulates events from a temporal duration of \SI{10}{\milli\second}, which signifies the voxel granularity.

\subsection{Comparison to State of the Art}
To validate the effectiveness of our model, we compared it with state-of-the-art trackers including event-based and image-based models. To facilitate equitable utilization of the high-temporal-resolution event data for image-based trackers, we constructed event images at a frame rate of $100$ FPS from the underlying high-temporal-resolution event data, aligning with the voxel granularity of \SI{10}{\milli\second} employed by our tracker. This strategy enabled the image-based comparative trackers to update tracklets at a heightened frequency of $100$ Hz, thereby enhancing their ability to track heavily occluded objects. This approach was designed to ensure that our model did not gain any inequitable advantage from access to high-temporal-resolution event data. All the results are obtained by the official implementation code to ensure a fair comparison, and performances are evaluated by the official metrics code. Our experiments are implemented on 2 RTX3090 GPUs and 128G RAM.


\begin{table*}[!h]
  \centering
  \caption{State-of-the-art Comparison on DSEC-MOT in Terms of Main Stream MOT Metrics.}
  \begin{tabular}{@{} l l r r r r r r r r r @{}}
    \hline
    Method & Sequence & HOTA$\uparrow$ & DetA$\uparrow$ & AssA$\uparrow$ & MOTA$\uparrow$ & IDF1$\uparrow$  & FN$\downarrow$ & FP$\downarrow$ & IDSw$\downarrow$ \\
    \hline
    \multirow{3}*{GTR \cite{zhou2022global}}
    & interlaken\_00\_d     & 55.7 & 65.1 & 47.7 & 72.6 & 64.5 &   610 &    93 &  5 \\
    & zurich\_city\_00\_b   & 33.5 & 34.0 & 33.3 & 35.1 & 40.1 & 2,835 &   481 & 87 \\
    & Combined              & 41.9 & 43.8 & 40.4 & 47.5 & 49.0 & 3,445 &   574 & 92 \\
    \hline
    \multirow{3}*{SiamMOT \cite{shuai2021siammot}}
    & interlaken\_00\_d     & 60.1 & 69.3 & 52.1 & 77.3 & 67.0 &   363 &   215 &  9 \\
    & zurich\_city\_00\_b   & 44.0 & 39.1 & 50.6 & 28.1 & 55.4 & 1,788 & 1,893 & 85 \\
    & Combined              & 49.1 & 47.4 & 51.4 & 44.4 & 59.1 & \textbf{2,151} & 2,108  & 94 \\
    \hline
    \multirow{3}*{ByteTrack \cite{zhang2022bytetrack}}
    & interlaken\_00\_d     & 58.6 & 67.5 & 50.9 & 75.6 & 66.3 &    583 &    42  &  5 \\
    & zurich\_city\_00\_b   & 36.3 & 30.1 & 43.8 & 34.5 & 47.0 &  3,279 &   146  &  9 \\
    & Combined              & 44.9 & 42.3 & 47.9 & 48.1 & 54.5 &  3,862 &   188  & 14 \\
    \hline
    \multirow{3}*{Trackformer \cite{meinhardt2022trackformer}}
    & interlaken\_00\_d     & 52.6 & 64.4 & 43.1 & 73.4 & 62.2 &    480 &   198  & 11 \\
    & zurich\_city\_00\_b   & 42.5 & 39.9 & 45.8 & 36.4 & 53.2 &  2,125 & 1,156  & 50 \\
    & Combined              & 45.9 & 47.2 & 45.0 & 48.6 & 56.2 &  2,605 & 1,354  & 61 \\    
    \hline
    \multirow{3}*{Ours}
    & interlaken\_00\_d     & 62.3 & 72.5 & 53.6 & 78.6 & 66.7 &   543  &   10   &  1 \\
    & zurich\_city\_00\_b   & 47.0 & 38.6 & 57.5 & 42.8 & 60.6 &  2,857 &  137   &  6 \\
    & Combined              & \textbf{52.5} & \textbf{49.5} & \textbf{55.7} & \textbf{54.7} & \textbf{62.9} &   3,400 & \textbf{147} & \textbf{7} \\
    \hline
  \end{tabular}
  \label{tab:res1}
\end{table*}

\begin{figure*}[!h]
    \includegraphics[width=\linewidth]{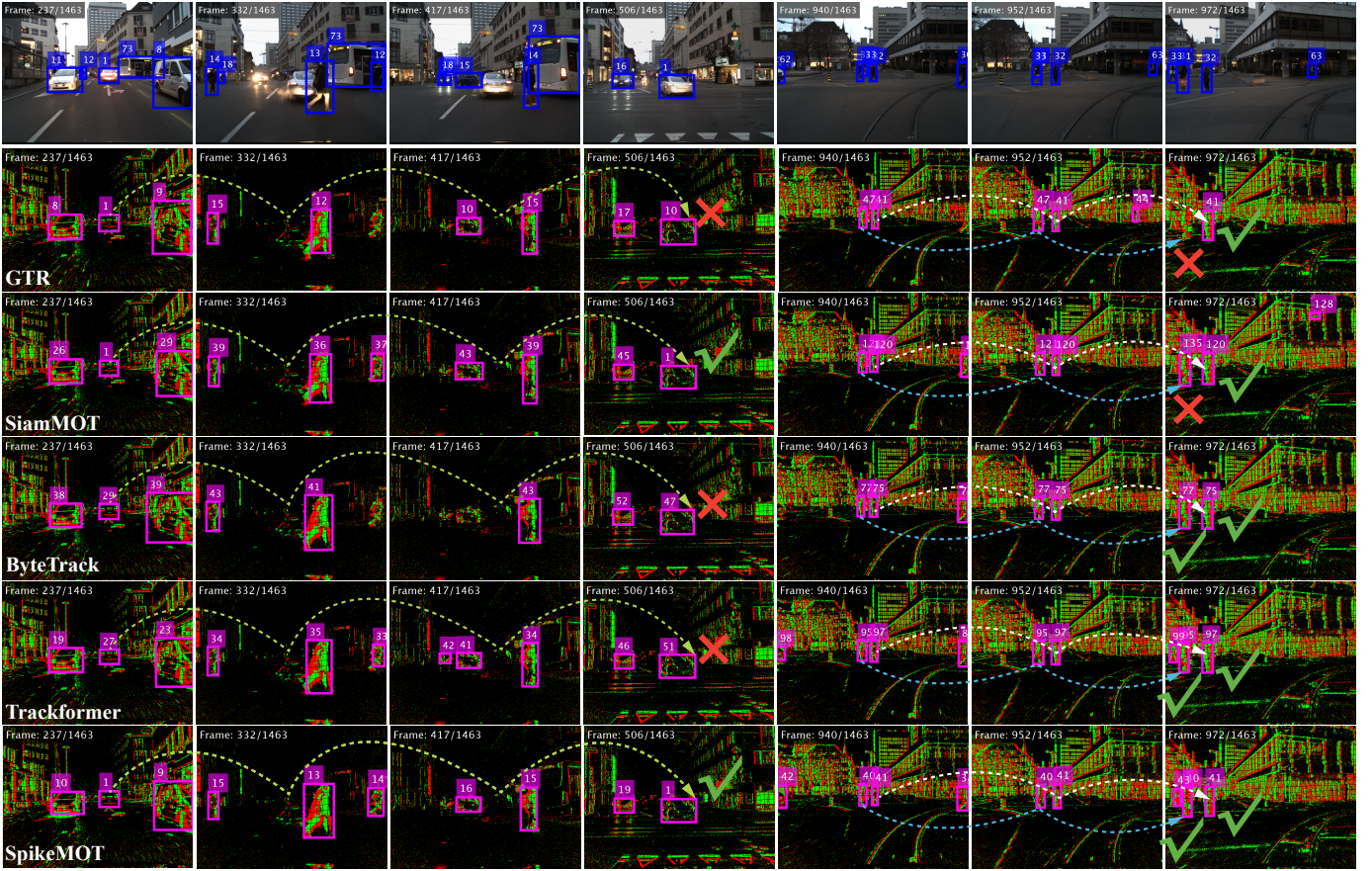}
    \centering
    \caption{Qualitative comparison of \name with state-of-the-art trackers under two scenarios. The left scenario (the first 4 columns) involves a busy intersection where the car in the middle is occluded by pedestrians, while the right scenario (the following 3 columns) features multiple persons moving together. The first row shows the groundtruth bounding boxes of the event domain showcased in the image domain, and the subsequent rows exhibit the trackers’ performance. In the left scene of the second row, the GTR tracker lost Car 1, which swapped identity with Car 10. In the right scene, GTR correctly tracked Person 41, but lost track of Person 47. In the third row, SiamMOT successfully tracked Car 1 after it was occluded, but an incorrect identity was assigned to Person 121, leading to a swap with Person 135. The fourth row demonstrates that ByteTrack was unable to track Car 29, while accurately tracking Pedestrians 75 and 77 in the corresponding scene on the right. Similarly, in the fifth row, Trackformer could not track the designated vehicle, but it was successful in tracking the specified pedestrians in the corresponding scene on the right. Finally, in the sixth row, our \name exhibited successful tracking of Car 1 that was occluded at the intersection, and accurately tracked the group of individuals moving together.}
    \label{fig:qual}
\end{figure*}

\begin{table*}[!h]
  \centering
  \caption{State-of-the-art Comparison on FE240hz in Terms of Main Stream MOT Metrics.}
  \begin{tabular}{@{} l l r r r r r r r r @{}}
    \hline
       Method & Sequence & HOTA$\uparrow$ & DetA$\uparrow$ & AssA$\uparrow$ & MOTA$\uparrow$ & IDF1$\uparrow$ & FP$\downarrow$ & FN$\downarrow$ & IDSw$\downarrow$ \\
    \hline
    \multirow{4}*{GTR \cite{zhou2022global}}
     & airplane         & 67.7 & 67.7 & 67.7 & 89.8 & 94.8 &     217  &   123  & 0 \\
     & bike\_hdr        & 64.0 & 64.0 & 64.0 & 82.0 & 90.8 &     418  &   297  & 0 \\    
     & dog\_motion      & 65.9 & 65.9 & 65.9 & 83.9 & 91.5 &     541  &   131  & 0 \\
     & elephant\_motion & 55.3 & 55.3 & 55.3 & 74.1 & 85.9 &     907  &   205  & 0 \\
     & Combined         & 63.2 & 62.8 & 63.7 & 82.0 & 90.6 &   2,083  &   756  & 0 \\
     \hline
    \multirow{4}*{SiamMOT \cite{shuai2021siammot}}
     & airplane         & 78.4 & 78.4 & 78.4 & 97.1 & 98.5 &      49  &    49  & 0 \\
     & bike\_hdr        & 70.8 & 70.8 & 70.8 & 86.6 & 93.3 &     270  &   261  & 0 \\    
     & dog\_motion      & 75.5 & 75.5 & 75.5 & 94.9 & 97.4 &     122  &    90  & 0 \\
     & elephant\_motion & 69.6 & 69.6 & 69.6 & 91.6 & 95.8 &     201  &   160  & 0 \\
     & Combined         & 73.5 & 73.1 & 73.8 & 92.4 & 96.2 &     642  &   560  & 0 \\
     \hline
    \multirow{4}*{ByteTrack \cite{zhang2022bytetrack}}
     & airplane         & 60.1 & 60.1 & 60.1 & 76.7 & 87.4 &      639 &   139 & 0 \\
     & bike\_hdr        & 73.8 & 73.8 & 73.8 & 92.9 & 96.4 &      158 &   125 & 0 \\    
     & dog\_motion      & 76.3 & 76.3 & 76.3 & 96.2 & 98.1 &      106 &    53 & 0 \\
     & elephant\_motion & 54.8 & 54.8 & 54.8 & 72.4 & 84.7 &      998 &   188 & 0 \\
     & Combined         & 67.1 & 66.2 & 68.0 & 84.8 & 92.0 &    1,901 &   505 & 0 \\
     \hline
    \multirow{4}*{Trackformer \cite{meinhardt2022trackformer}}
     & airplane         & 73.7 & 73.7 & 73.7 & 96.5 & 98.3 &       58 &   58 & 0 \\
     & bike\_hdr        & 77.7 & 77.7 & 77.7 & 98.4 & 99.2 &       32 &   32 & 0 \\    
     & dog\_motion      & 75.0 & 75.0 & 75.0 & 95.8 & 97.9 &       87 &   87 & 0 \\
     & elephant\_motion & 69.4 & 69.4 & 69.4 & 93.5 & 96.7 &      141 &  140 & 0 \\
     & Combined         & 74.1 & 73.7 & 74.4 & 96.0 & 98.0 &      318 &  317 & 0 \\
     \hline
    \multirow{4}*{STNet \cite{zhang2022spiking}}
     & airplane         & 61.9 & 61.9 & 61.9 & 78.2  & 89.1 &      364 &   364 & 0 \\
     & bike\_hdr        & 69.8 & 69.8 & 69.8 & 95.6  & 97.8 &       88 &    88 & 0 \\    
     & dog\_motion      & 60.0 & 60.0 & 60.0 & 77.0  & 88.5 &      481 &   481 & 0 \\
     & elephant\_motion & 62.6 & 62.6 & 62.6 & 83.5  & 91.7 &      355 &   355 & 0 \\
     & Combined         & 63.7 & 63.4 & 64.0 & 83.7  & 91.8 &    1,288 & 1,288 & 0 \\  
    \hline
    \multirow{4}*{Ours}
     & airplane         & 86.4 & 86.4 & 86.4 & 99.8 & 99.9 &          4 &   4 & 0 \\
     & bike\_hdr        & 88.1 & 88.1 & 88.1 & 99.6 & 99.8 &         16 &   0 & 0 \\    
     & dog\_motion      & 89.0 & 89.0 & 89.0 & 98.4 & 99.2 &         39 &  27 & 0 \\
     & elephant\_motion & 86.0 & 86.0 & 86.0 & 98.6 & 99.3 &         34 &  26 & 0 \\
     & Combined         & \textbf{87.4} & \textbf{87.4} & \textbf{87.5} & \textbf{99.0} & \textbf{99.5} &         \textbf{93} &  \textbf{57} & 0 \\
    \hline
  \end{tabular}
  \label{tab:res2}
\end{table*}

\begin{figure*}[!h]
    \includegraphics[width=0.95\linewidth]{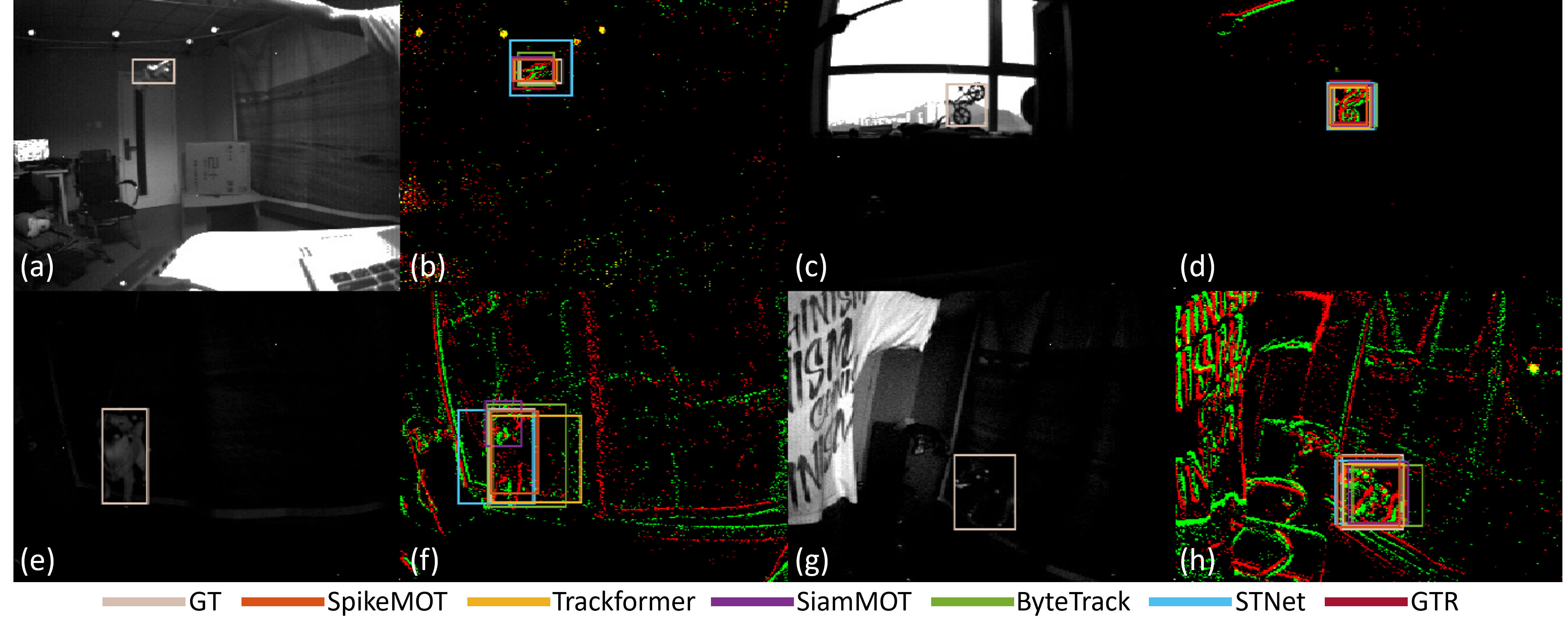}
    \centering
    \caption{Qualitative comparison of \name with state-of-the-art trackers under FE240hz dataset. The first row displays the \textit{airplane} and \textit{bike\_hdr} sequences, while the second row exhibits the \textit{dog\_motion} and \textit{elephant\_motion} sequences. Panels (a), (c), (e), and (g) showcase the groundtruth within the gray images corresponding to the associated event frames (b), (d), (f), and (h). Our approach effectively maintained a secure lock on the target under the degraded conditions.}
    \label{fig:qual_fe240hz}
\end{figure*}

\subsubsection{DSEC-MOT Results}
The DSEC-MOT testing sequences are characterized by considerable occlusions, making the dataset an ideal benchmark for evaluating the target association performance of tracking methods. We chose state-of-the-art CNN-based and Transformer-based trackers as baselines to evaluate the effectiveness of \name. Our model demonstrates the best performance across a majority of evaluation metrics (\cref{tab:res1}). Notably, \name has showcased a significant reduction in identity switch occurrence compared to other tracking methods, indicating its competence in effectively addressing the challenges of object association under heavy occlusions in DSEC-MOT. These findings underscore the discriminative sparse spatiotemporal features extracted by our model. In \cref{fig:qual}, we present a qualitative comparison of \name alongside state-of-the-art trackers in two distinct scenarios within the DSEC-MOT dataset. This visual representation illustrates our approach's capability to accurately track targets in situations of complete occlusion.

\subsubsection{FE240hz Results}
We specifically selected from the FE240hz dataset the testing sequences that involved high-speed targets and exhibited severe camera motion. The significant camera vibrations in these sequences result in the generation of a substantial volume of background events. The merging of rapidly-changing background events with fast-moving target events presents the challenges aimed to evaluate the trackers' proficiency in locking targets under such demanding conditions. The performance of our \name is illustrated in \cref{tab:res2}, which demonstrates a considerable advantage over state-of-the-art tracking methods across all metrics. \cref{fig:qual_fe240hz} offers a qualitative comparison of \name with state-of-the-art trackers within the FE240hz dataset. These results confirm \name's reliable target-locking abilities driven by the sparse motion features, especially in challenging scenarios involving fast-moving objects amidst a substantial background event presence. 

\subsubsection{Runtime Comparison}
\begin{figure}[htbp]
    \includegraphics[width=0.9\linewidth]{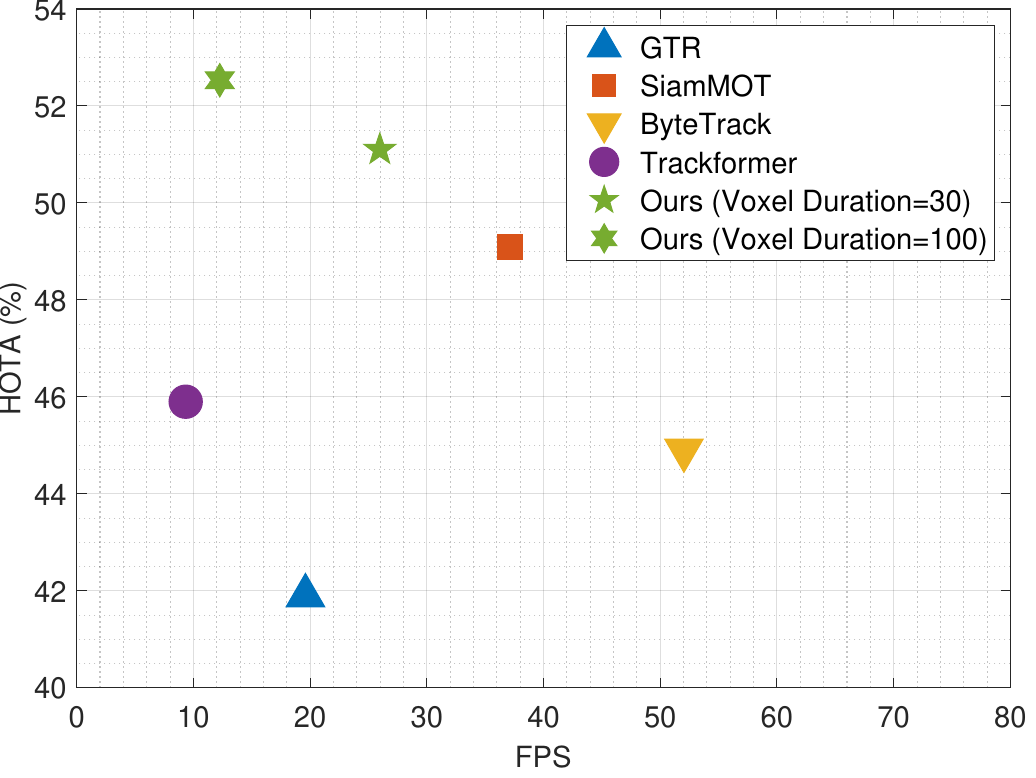}
    \centering
    \caption{Runtime comparison regarding to the HOTA metric on the DSEC-MOT testing dataset. With a voxel duration of $30$, our approach achieved a speed of $26$ FPS and an HOTA of $51.1$. Employing a voxel duration of $100$ led to a speed of $12$ FPS and a corresponding HOTA of $52.5$.}
    \label{fig:fps}
\end{figure}

We present a comparison of state-of-the-art trackers with respect to tracking speed and tracking proficiency. All trackers were evaluated on the DSEC-MOT testing dataset at the resolution of 640$\times$480 and executed on the machine with an RTX3090 GPU and an AMD EPYC 24-core CPU. \cref{fig:fps} illustrates the runtime and the corresponding HOTA metric for all trackers. With the voxel duration of $100$, our approach operated at the speed of $12$ FPS and attained the HOTA of $52.5$. When using a voxel duration of $30$, our method achieved a speed of $26$ FPS with the HOTA of $51.1$. The influence of voxel duration on tracking speed is analyzed in Section 5.5.3.

\subsection{Ablations}
\subsubsection{Recurrent Pathways with Sparse Features} 
Through the analysis of various network models, we have observed that the integration of recurrent pathways and sparse features leads to enhanced tracking performance. Particularly, SNNs, which have intra-neuron recurrent connections, such as the self-recurrences in response and refractory kernels and the inhibition process, exhibit remarkable performance \cite{he2020comparing}. \cref{tab:abaltion} illustrates the comparison of three distinct backbone network types. The first type lacks recurrent pathways and produces dense features, exemplified by CNN. The second type incorporates recurrent pathways and generates dense features, characterized by LSTM. The third type integrates recurrent pathways and yields sparse features, represented by SNNs. This investigation revealed that the SNN employing SRM neurons outperformed its counterparts across all metrics. Thus, this particular configuration emerged as the top-performing option in our study.

\begin{table}[!h]
  \centering
  \caption{Comparison Among Three Categories of Backbone Networks and the Configuration Utilizing SNN with SRM Neurons Emerged as the Top-performing Across All Metrics.}
  \begin{tabular}{@{} c c c c c c @{}}
    \hline
     Backbone network & HOTA$\uparrow$ & DetA$\uparrow$ & AssA$\uparrow$ & MOTA$\uparrow$ & IDF1$\uparrow$ \\
    \hline
    CNN                               & 46.2 & 44.0 & 48.6 & 50.1 & 59.8 \\
    LSTM                              & 51.1 & 49.4 & 52.9 & 54.3 & 60.2 \\
    SNN (LIF neuron)                  & 51.2 & 49.3 & 53.2 & 54.4 & 60.4 \\
    \rowcolor{GRAY} SNN (SRM neuron)  & 52.5 & 49.5 & 55.7 & 54.7 & 62.9 \\
    \hline
  \end{tabular}
  \label{tab:abaltion}
\end{table}

Our analysis identified two primary factors contributing to this particular configuration. The first factor is the inherent sparsity of SNN features. At any given time, only a small fraction of neurons in the SNN are active. This sparse activation pattern allows the SNN to concentrate exclusively on relevant information while disregarding extraneous data. This focused attention discriminates against semantic distractors and contributes to the tracking efficacy. The second factor is the presence of intra-neuron recurrent connections within the SNN, such as the self-recurrences in response and refractory kernels and the inhibition process. These recurrent pathways enable the network to maintain temporal information for an extended period. By memorizing longer-term information, the network can establish better associations between objects, resulting in enhanced tracking accuracy. Moreover, the self-recurrence within each neuron significantly reduces the number of trainable parameters within the SNN, leading to a more compact model and improved generalization abilities \cite{he2020comparing}. 

In addition, the superiority of the SNNs utilizing SRM neurons over those using LIF neurons lies in their ability to retain information and exhibit diverse temporal dynamics over time. Unlike LIF neurons, which hard reset the membrane potential to the resting potential upon reaching the threshold, SRM neurons possess a unique soft decay mechanism that the membrane potential is partially inhibited and the inhibition gradually decays over time (\cref{fig:siam}c). This characteristic enables the SNN with SRM neurons to memorize longer-term dependencies and thereby improve tracking performance. Besides, if the partially inhibited membrane potential persists above the threshold, it will trigger a series of successive spikes. This spiking burst, absent in the LIF neuron output, showcases the SRM neurons' capability of generating richer spiking patterns. Such patterns offer a more detailed and discriminative representation, allowing for better differentiation between similar and dissimilar patterns as well as greater resilience to noisy backgrounds.

\subsubsection{Neuron's Threshold Potential}
\begin{figure}[htbp]
    \includegraphics[width=0.7\linewidth]{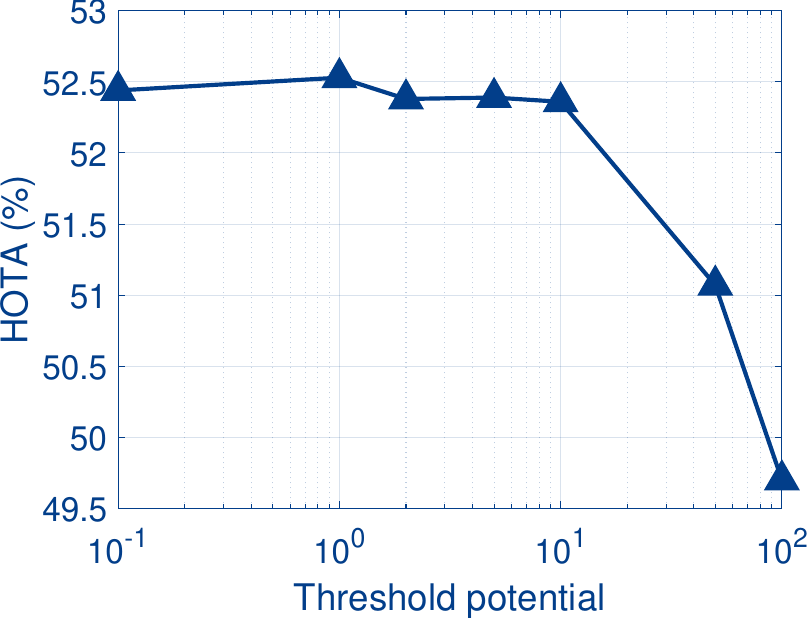}
    \centering
    \caption{The impact of neuron’s threshold potential on tracking performance. The effect is negligible when the threshold potential varies between 0.1 and 10.0, but the tracking accuracy is considerably hampered when the threshold potential exceeds 10.0.}
    \label{fig:th_poten}
\end{figure}

The threshold potential determines the firing behavior of neurons and thus we examined its impact on the tracking accuracy using the DSEC-MOT testing set. \cref{fig:th_poten} reveals that the influence is minimal when the threshold potential lies between 0.1 and 10.0. In our SNN model, a synapse network (CNN) precedes the spiking neurons. During the training process, the output of the synapse network is optimized by tuning the synaptic weights. This allows the output to accommodate the spiking neurons' threshold potentials, enabling the SNN model to produce the spikes that maximize tracking performance. However, the network’s adaptation achieved solely through synaptic plasticity still has certain limitations. If the threshold potential is set too high, neurons may still fire less frequently, even with adjusted synaptic weights, leading to feature loss and reduced accuracy.

\subsubsection{Voxel Duration and Granularity of Motion Features}
\begin{figure}[htbp]
    \centering
    \subfloat{\includegraphics[width=0.51\linewidth]{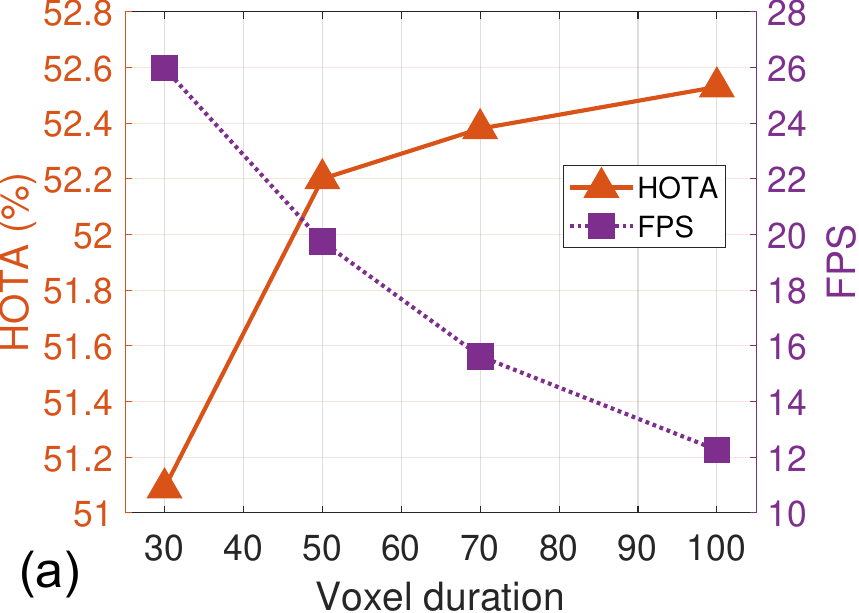}}
    \hfill
    \subfloat{\includegraphics[width=0.46\linewidth]{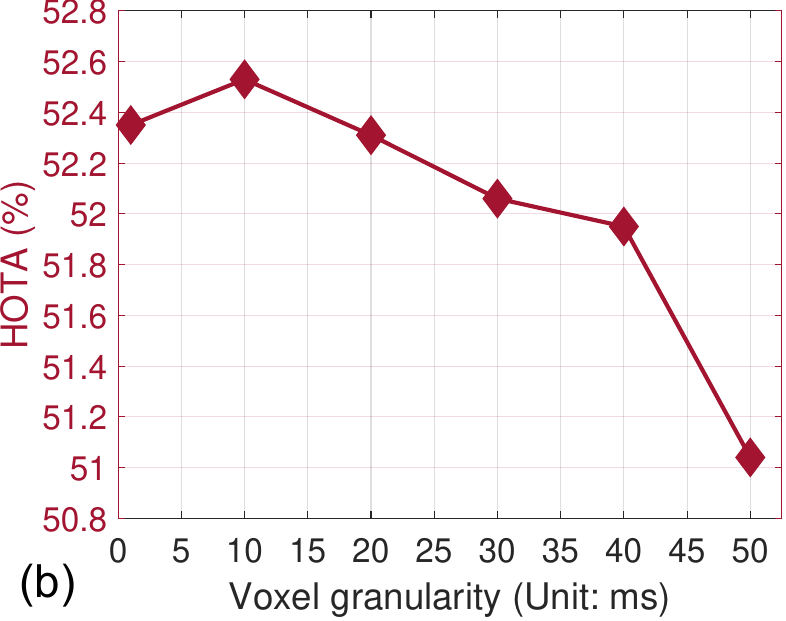}}
    \caption{The influence of voxel duration and voxel granularity on tracking performance as well as the impact of voxel duration on tracking speed. (a): A longer voxel duration includes greater motion information, resulting in improved tracking outcomes. However, this enhancement comes at the cost of increased computational load, leading to a decrease in tracking speed. (b): A smaller voxel granularity yields a finer resolution of motion details, enhancing the tracking effectiveness. Nevertheless, excessively small voxel granularity leads to a reduction in the voxel time length, which hampers tracking proficiency.}
    \label{fig:time}
\end{figure}

\cref{fig:time} evaluates the impact of voxel duration and voxel granularity on tracking outcomes, and it reveals that optimal tracking results are achieved when longer voxel durations and smaller voxel granularities are employed. The voxel duration represents the length of the time window over which the spatiotemporal features are computed. Longer voxel durations capture more temporal context and provide a richer representation of object motion, helping to handle occlusions. Notably, increased voxel durations also augment computational complexity, which in turn curtails tracking speed. On the other hand, voxel granularity refers to the sampling period for spatiotemporal features. A finer voxel granularity enhances the resolution of motion details within the spatiotemporal features, assisting in capturing subtle movements and rapid changes. However, given a fixed voxel dimension, excessively fine voxel granularities shorten the time length of the voxel, leading to a decline in tracking effectiveness.

\subsubsection{Detector's Impact on Tracking Outcomes}
\begin{figure}[!h]
    \centering
    \subfloat{\includegraphics[width=0.48\linewidth]{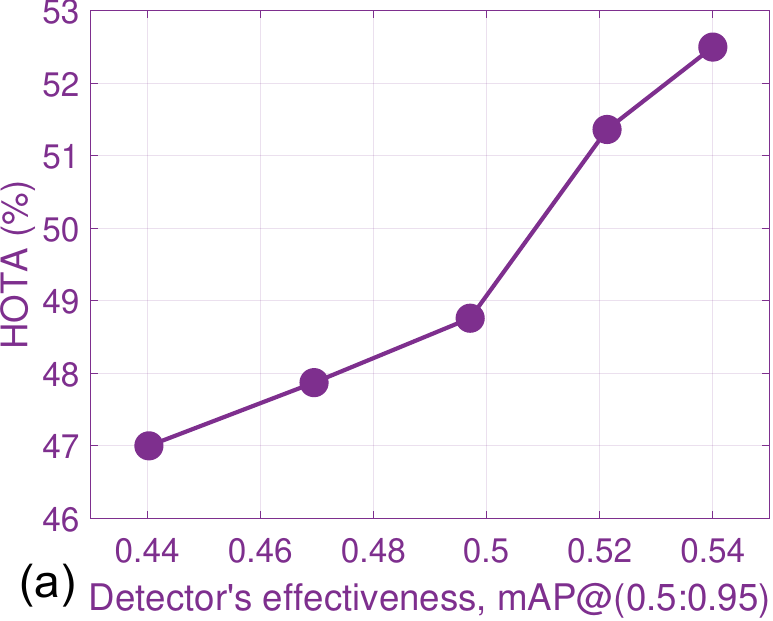}}
    \hfill
    \subfloat{\includegraphics[width=0.48\linewidth]{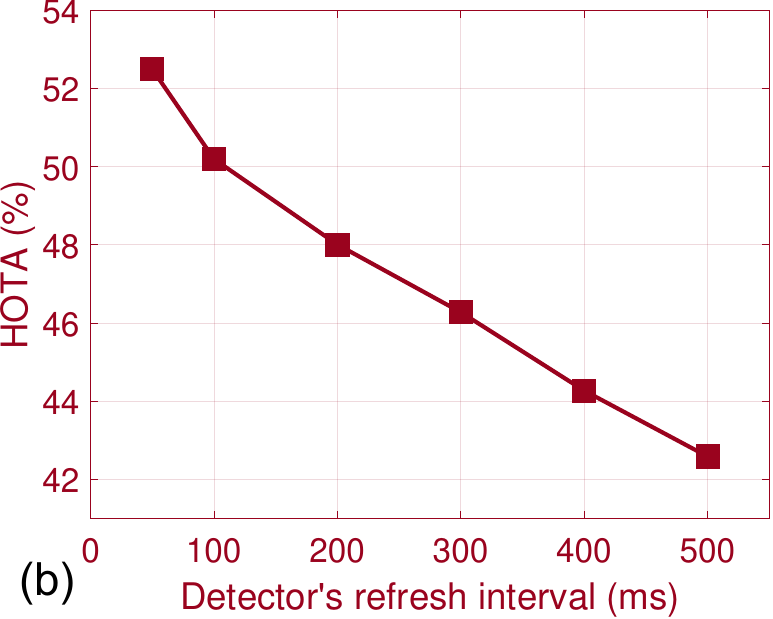}}
    \caption{Detector's impact on tracking outcomes. (a): A positive correlation between the detector's mAP@(0.5:0.95) and the tracking accuracy. (b): A negative correlation between the detector's refresh interval and the tracking accuracy.}
    \label{fig:yolo_qual}
\end{figure}

We investigated the impact of detector efficacy on tracking accuracy (\cref{fig:yolo_qual}a). By manipulating the number of training epochs, we obtained the detectors with varying levels of effectiveness. These detectors were then evaluated for their influence on tracking performance under the DSEC-MOT testing set. Results indicated that the model incorporating a more effective detector exhibited a higher tracking accuracy, due to the reduced false positives and false negatives. Moreover, templates updated by an effective detector tend to represent the target objects more accurately, resulting in improved tracking outcomes. Therefore, detectors with higher efficacy generally exert a greater positive impact on tracking performance.

We further examined the impact of detector refresh interval on tracking accuracy (\cref{fig:yolo_qual}b). Our experiments on the DSEC-MOT testing set revealed a negative correlation between the two. Specifically, a shorter refresh interval allows the detector to capture new objects more quickly and handle occlusions more effectively by rapidly re-detecting the object once it reappears. This is particularly advantageous in crowded scenarios or situations where objects frequently occlude one another. In addition, the decreased refresh interval helps the tracking system adapt to high object dynamics and appearance changes. When objects exhibit complex motion patterns or experience significant appearance changes over time, due to variations in perspective and distance, frequently running the detector enables the tracker to promptly update templates and maintain accurate representations of the objects.

\subsection{Limitation}
To achieve optimal tracking efficiency and proficiency, parameters could be customized based on specific scenarios. For instance, in scenarios with minimal occlusions, a shorter voxel duration is adequate, whereas employing a longer voxel duration would impose an unnecessary computational burden. Additionally, in relatively stable scenarios characterized by the absence of new objects and minimal appearance changes in existing objects over time, a shorter detector refresh interval is unnecessary and would introduce an excessive computational load. Consequently, the incorporation of a self-adaptive mechanism for parameter selection tailored to specific scenarios can yield advantageous outcomes in terms of tracking accuracy and computational efficiency.

\section{Conclusion}
In this paper, we have introduced \name, a novel event-based tracking method that integrates the SNNs employing SRM neurons into a Siamese framework for MOT in the event domain. Additionally, we have presented DSEC-MOT, the first real-world event-based MOT benchmark that features fine-grained annotations and contains severe occlusions, extensive trajectory intersections, uniform appearance, and objects out of view. Experimental results reveal that \name achieves state-of-the-art tracking accuracy across DSEC-MOT and FE240hz, despite facing formidable real-world scenarios. We conducted an analysis of three backbone network variants, revealing that the SNNs employing SRM neurons outperformed their counterparts across all metrics. This success can be attributed to the recurrent pathways and the sparse motion features. We further investigated the impact of neuron's threshold potential on the tracking accuracy, finding that the threshold potential ranging from 0.1 to 10.0 has limited influence, due to the adaption of the synaptic network. We also performed extensive analyses to investigate the effects of motion feature duration and granularity on tracking outcomes, discovering that extended durations and reduced granularities contribute positively to tracking outcomes. Moreover, we explored the influence of detectors on tracking results, identifying that detectors with higher efficacy and shorter refresh intervals enhance tracking performance. In summary, our proposed \name and DSEC-MOT provide a robust benchmark for future research. Going forward, we plan to investigate further into SNN architectures and their potential applications in event-based vision tasks.

\section*{Acknowledgements}
This work was supported in part by the Hong Kong Research Grants Council under the Research Impact Fund project R7003-21. This work was also supported by the AI Chip Center for Emerging Smart Systems (ACCESS), sponsored by InnoHK funding, Hong Kong SAR.

\bibliographystyle{IEEEtran}
\bibliography{IEEEabrv,egbib}

\begin{IEEEbiography}[{\includegraphics[width=1in,height=1.25in]{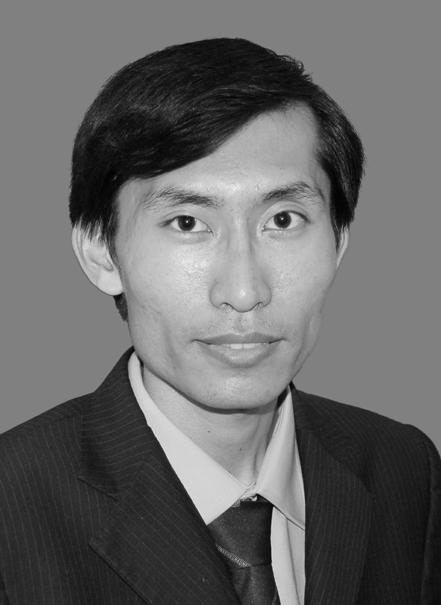}}]{Song Wang}
received his B.E. from Harbin Institute of Technology, China, and his M.Phil. from The University of Hong Kong. He is currently a Ph.D. student in the Department of Electrical and Electronic Engineering, The University of Hong Kong. His research interests include computer vision and computer architecture.
\end{IEEEbiography}

\begin{IEEEbiography}[{\includegraphics[width=1in,height=1.25in]{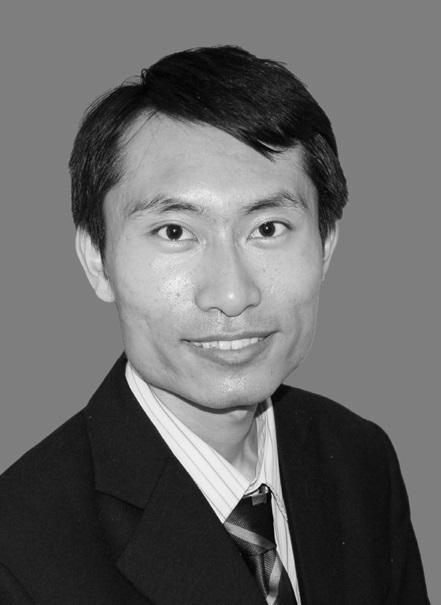}}]{Zhu Wang}
received his B.E. from Harbin Institute of Technology, China, and his M.Phil. from City University of Hong Kong. He is currently a Ph.D. student in the Department of Electrical and Electronic Engineering, The University of Hong Kong. His research interests include neuromorphic computing, A.I. accelerators, and VLSI design.
\end{IEEEbiography}

\begin{IEEEbiography}[{\includegraphics[width=1in,height=1.25in]{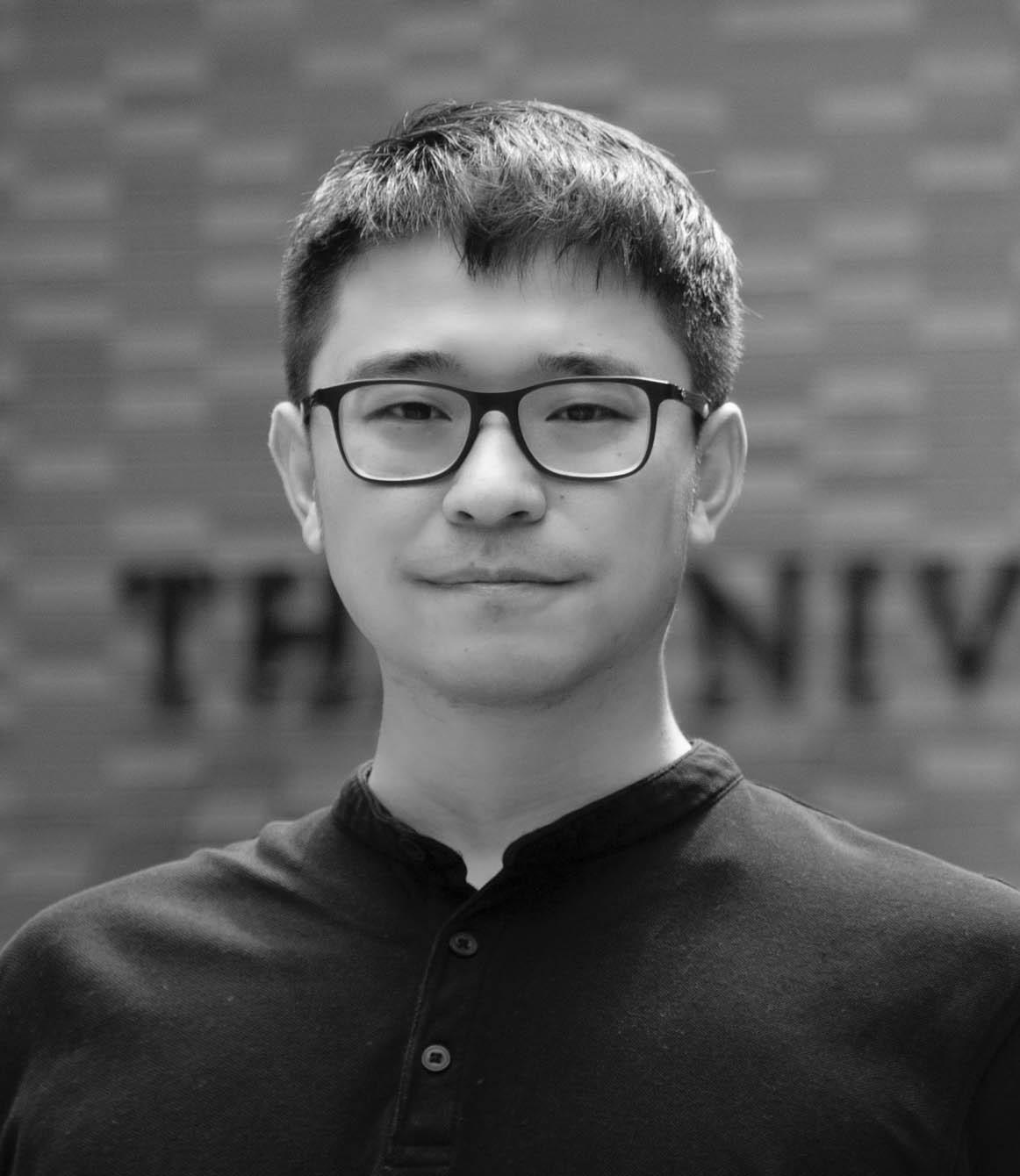}}]{Can Li}
is an Assistant Professor in the Department of Electrical and Electronic Engineering, The University of Hong Kong. Previously, he was a postdoctoral researcher at Hewlett Packard Labs in Palo Alto, California. He obtained his B.S. (2009) and M.S. (2012) from Peking University and his Ph.D. (2018) from the University of Massachusetts, Amherst. His research interests include neuromorphic computing, A.I. accelerators, and non-volatile memories. He has received several awards, such as RGC Early Career Awards (HKSAR), 2021, NSFC Excellent Young Scientists Fund, 2021, and Croucher Tak Wah Mak Innovation Award, 2023. He has served as a technical program committee member for prestigious events such as DAC 2022, DAC 2023, ICCAD 2023, ASP-DAC 2023, and ASP-DAC 2024. He is also an editorial advisory board member of APL Machine Learning and a reviewer for high-impact journals such as Nature Electronics, Nature Communications, and Science Advances.
\end{IEEEbiography}

\begin{IEEEbiography}[{\includegraphics[width=1in,height=1.25in]{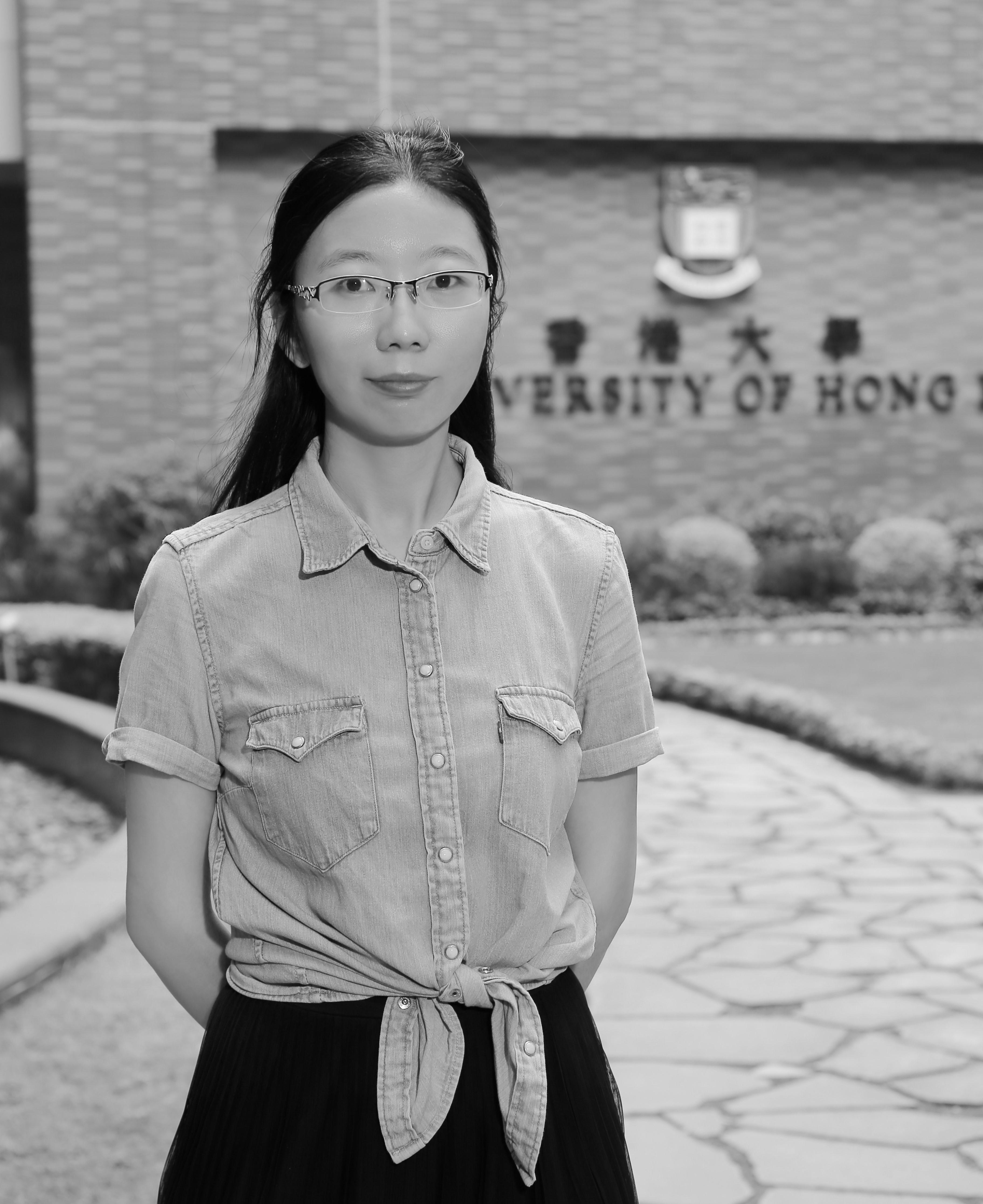}}]{Xiaojuan Qi}
is an Assistant Professor in the Department of Electrical and Electronic Engineering, The University of Hong Kong. Previously, she was a postdoctoral researcher in the Department of Engineering Science, University of Oxford. She obtained her Ph.D. in Computer Science from the Chinese University of Hong Kong in 2018 and her B.E. in Electronic Science and Technology from Shanghai Jiao Tong University in 2014. From Sept. 2016 to Nov. 2016, she was a visiting student in the Machine Learning Group, University of Toronto. She has carried out an internship at Intel Intelligent Systems Lab from May 2017 to Nov. 2017. She has won several awards such as first place in ImageNet Semantic Parsing Challenge, the Hong Kong Ph.D. Fellowship Award, Doctoral Consortium Travel Award in CVPR’18, and Outstanding Reviewer Award in ICCV’17 and ICCV’19. She was an area chair for AAAI 2021, CVPR 2021, ICCV 2021, AAAI 2022, AAAI 2023, WACV 2023, CVPR 2023, and NeurIPS 2023.
\end{IEEEbiography}

\begin{IEEEbiography}[{\includegraphics[width=1in,height=1.25in]{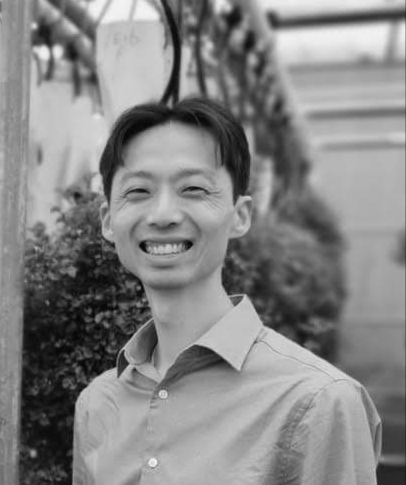}}]{Hayden Kwok-Hay So}
(M’03-SM’15) received his B.S., M.S., and Ph.D. in electrical engineering and computer sciences from the University of California, Berkeley in 1998, 2000, and 2007, respectively. He is currently an Associate Professor and co-director of the computer engineering program at the Department of Electrical and Electronic Engineering, The University of Hong Kong. He is also the co-director and founding member of the Joint Lab on Future Cities at The University of Hong Kong. His research focuses on highly efficient reconfigurable computing system designs and their applications. He has been awarded the CODES+ISSS Test-of-Time award, in 2021, the IEEE-HKN C. Holmes MacDonald Outstanding Teaching Award, in 2021, the Croucher Innovation Award, in 2013, and the University Outstanding Teaching Award in 2012. He is an active member of the international reconfigurable computing community, having served as organizer and technical program chairs of many international conferences in the field.

\end{IEEEbiography}

%
%
%
%
%



\enlargethispage{-5in}

\end{document}